\documentclass[10pt,journal,compsoc]{IEEEtran}
\ifCLASSOPTIONcompsoc
  \usepackage[nocompress]{cite}
\else
  \usepackage{cite}
\fi
\ifCLASSINFOpdf
\else
\fi

\usepackage{amsfonts}

\usepackage[noend]{algorithmic}

\usepackage{mathtools}
\usepackage{amsmath}
\usepackage{amsthm}
\usepackage{eucal}
\usepackage{amssymb}
\usepackage{mathrsfs}

\usepackage{textcomp} 
\usepackage{epstopdf}

\usepackage{adjustbox}

\usepackage{caption}
\usepackage{subcaption}

\usepackage{multirow,booktabs}
\usepackage[ruled,vlined]{algorithm2e}
\usepackage{verbatim}

\usepackage{url}
\usepackage{hyperref}

\DeclareMathOperator*{\argmin}{argmin}

\usepackage{xcolor,colortbl}

\def\R{\mathbb{R}}

\def\x{\mathbf{x}}
\def\c{\mathbf{c}}

\def\X{\mathbf{X}}

\def\C{\mathbf{C}}

\def\R{\mathbf{R}}

\def\I{\mathbf{I}}

\begin{document}
%
\title{Scalable Initialization Methods for \\Large-Scale Clustering}
%
%
%
%

\author{Joonas~{H{\"a}m{\"a}l{\"a}inen},
        Tommi~{K{\"a}rkk{\"a}inen},
        and~Tuomo {Rossi}
\IEEEcompsocitemizethanks{\IEEEcompsocthanksitem J. {H{\"a}m{\"a}l{\"a}inen}, T. {K{\"a}rkk{\"a}inen} and T. Rossi are with the Faculty of Information Technology, University of Jyvaskyla, Finland,
FI-40014.} 
\protect\\%
E-mail: joonas.k.hamalainen@jyu.fi
\thanks{This work has been submitted to the IEEE for possible publication. Copyright may be transferred without notice, after which this version may no longer be accessible.}}

%
%

\markboth{Preprint submitted to IEEE Transactions on Big Data}%
{Shell \MakeLowercase{\textit{et al.}}: Bare Demo of IEEEtran.cls for Computer Society Journals}
%



\IEEEtitleabstractindextext{%
\begin{abstract}
In this work, two new initialization methods for K-means clustering are proposed. Both proposals are based on applying a divide-and-conquer approach for the K-means$\parallel$ type of an initialization strategy. The second proposal also utilizes multiple lower-dimensional subspaces produced by the random projection method for the initialization. The proposed methods are scalable and can be run in parallel, which make them suitable for initializing large-scale problems. In the experiments, comparison of the proposed methods to the K-means++ and K-means$\parallel$ methods is conducted using an extensive set of reference and synthetic large-scale datasets. Concerning the latter, a novel high-dimensional clustering data generation algorithm is given. The experiments show that the proposed methods compare favorably to the state-of-the-art. We also observe that the currently most popular K-means++ initialization behaves like the random one in the very high-dimensional cases.
\end{abstract}

}

\maketitle

\IEEEdisplaynontitleabstractindextext

%
\IEEEpeerreviewmaketitle

\IEEEraisesectionheading{\section{Introduction}\label{sec:introduction}}

%
%
%
%

\IEEEPARstart{C}{lustering} is one of the core techniques in data mining. Its purpose is to form groups from data in a way that the observations within one group, the cluster, are similar to each other and dissimilar to observations in other groups. Prototype-based clustering algorithms, such as the popular K-means \cite{Llo1982}, are known to be sensitive to initialization \cite{Emr2012}, i.e., the selection of initial prototypes. A proper set of initial prototypes can improve the clustering result and decrease the number of iterations needed for the convergence of an algorithm \cite{hamalainen2017comparison}. The initialization of K-means was remarkably improved by the work of Arthur and Vassilvitskii \cite{ArtVas2007}, where they proposed the K-means++ method. There, the initial prototypes are determined by favoring distinct prototypes, which in high probability are not similar to the already selected ones.

A drawback of K-means++ is that the initialization phase requires $K$ inherently sequential passes over the data, since the selection of a new initial prototype depends on the previously selected prototypes. Bahmani et al. \cite{Bah2012} proposed a parallel initialization method called K-means$\parallel$. The K-means$\parallel$ speeds up initialization by sampling each point independently and by updating sampling probabilities less frequently. Independent sampling of the points enables parallelization of the initialization, thus providing a speedup over K-means++. However, for example MapReduce based implementation of K-means$\parallel$ needs multiple MapReduce jobs for the initialization. The MapReduce K-means++ method \cite{Xu2014} tries to address this issue, as it uses one MapReduce job to select $K$ initial prototypes, which speeds up the initialization compared to K-means$\parallel$. Suggestions of parallelizing the second, search phase of K-means have been given in several papers (see, e.g., \cite{Dhi1999,Zha2009}). On a single machine, distance pruning approaches can be utilized to speed up K-means without affecting the clustering results \cite{elkan2003using,hamerly2010making}.

Dimension reduction has had an important role in making clustering algorithms more efficient. Over the years, various dimension reduction methods have been applied to decrease the dimension of data in order to speed up clustering algorithms \cite{Bou2010, FerBro2003, alzate2010multiway, napoleon2011new}. The key idea for improved efficiency is to solve an approximate solution to the clustering problem in a lower dimensional space. Dimension reduction methods are usually divided into two categories: feature selection methods and feature extraction methods \cite{Liu1998}. Feature selection methods aim to select a subset of the most relevant variables from the original variables. Correspondingly, feature extraction methods aim to transform the original dataset into a lower dimensional space while trying to preserve the characteristics (especially distances between the observations and the overall variability) of the original data.

A particular dimensional reduction approach for processing large datasets is the random projection (RP) method \cite{Ach2003}. Projecting data from the original space to a lower dimensional space while preserving the distances is the main characteristic of the RP method. This makes RP very appealing in clustering, whose core concept is dissimilarity. Moreover, classical dimension reduction methods such as the principal component analysis (PCA) \cite{Jol2002} become expensive to compute for high-dimensional spaces whereas RP remains computationally efficient \cite{BingMan2001}.

Fern and Brodley \cite{FerBro2003} proposed an ensemble clustering method based on RP. They showed empirically that aggregation of clustering results from multiple lower dimensional spaces produced by RP leads to better clustering results compared to a single clustering in lower dimensional space produced by PCA or RP. Other combinations of K-means and RP have been studied in several papers \cite{Coh2014,Bou2010,Bou2015,Car2012}. RP for K-means++ was analyzed in \cite{chan2017efficient}. Generally, the main idea is to create a lower dimensional dataset with RP and to solve the ensuing K-means clustering problem with less computational effort.

In general, K-means clustering procedure typically uses a non-deterministic initialization, such as K-means++, followed by the Lloyd's iterations \textemdash with multiple restarts. Prototypes corresponding to the smallest sum-of-squares clustering error are selected as the final clustering result. In \cite{HamKar2016}, such a multistart strategy was carried out during the initialization phase, thus reducing the need to repeat the whole clustering algorithm. More precisely, a parallel method based on K-means++ clustering of subsets produced by the distribution optimally balanced stratified cross-validation (DOB-SCV) algorithm \cite{MorenoTorres2012} was proposed and tested. Here, such an approach is developed further with the help of K-means$\parallel$ and RP. More precisely, we run K-means$\parallel$ method in a low-dimensional subsets created by RP. In contrast to the previous work \cite{HamKar2016}, the new methods also restrict the number of Lloyd's iterations in the subsets.

To summarize, the proposed initialization method reduces data processing with sampling, subsampling, and dimensional reduction, and then solves the K-means clustering problem in a coarse fashion. Moreover, from the perspective of parallel computing, using a parallelizable clustering method in the subset clustering allows fixing the number of subsets and treating each subset locally in parallel, hence improving the scalability. Therefore, the main purpose of this article is to propose two new algorithms for clustering initialization and compare them experimentally with K-means++ and K-means$\parallel$ using several large-scale datasets. For quantified testing and comparison of the methods, we also introduce a novel clustering problem generator for high dimension spaces (see Section \ref{Sec:KSPHERES}).

\section{Background}

In this section, we introduce the basic composition of the existing algorithms.

\subsection{K-means clustering problem and\newline \noindent\hangindent 0.8cm the basic algorithms}

Let $\X = \{\x_1,\x_2,...,\x_N\}$ be a dataset such that $\x_i \in \mathbb{R}^M \ \forall 1 \le i \le N$, and let $\C = \{\c_1,\c_2,...,\c_K\}$ be a set of prototypes, where each prototype also belongs to $\mathbb{R}^M$. The goal of the K-means clustering algorithm is to find a partition of $\X$ into $K$ disjoint subsets, by minimizing the sum-of-squares error (SSE) defined as
\begin{equation}
\label{SSE}
\text{SSE}(\C) = \sum_{\x \in \X} \min_{\c \in \C} {\|{{\c} - \x}\| }^2 .
\end{equation}
An approximate solution to the minimization problem with \eqref{SSE} is typically computed by utilizing the Lloyd's K-means algorithm \cite{Llo1982}. Its popularity is based on simplicity and scalability. Even if the cost function in \eqref{SSE} is mathematically nondifferentiable because of the $\min$-operator, it is easy to show that after the initialization, the K-means type of iterative relocation algorithm converges in finite many steps \cite{hamalainen2017comparison}. 

Prototype-based clustering algorithms, like K-means, are initialized before the prototype relocation (search) phase. The classical initialization algorithm, readily proposed in \cite{macqueen1967some}, is to randomly generate the initial set of prototypes. A slight refinement of this strategy is to select, instead of random points (from appropriate value ranges), random indices and use the corresponding observations in data as initialization \cite{forgy1965cluster}. Because of this choice, there cannot be empty clusters in the first iteration. Bradley and Fayyad \cite{bradley1998refining} proposed an initialization method where $J$ randomly selected subsets of the data are first clustered with K-means. Next, it forms a superset of the $J \times K$ prototypes obtained from the subset clustering. Finally, the initial prototypes are achieved as the result of K-means clustering of the superset.

Arthur and Vassilvitskii \cite{ArtVas2007} introduced the K-means++ algorithm, which improves the initialization of K-means clustering. The algorithm selects first prototype at random, and then the remaining $K-1$ prototypes are sampled using probabilities based on the squared distances to the already selected set, thus favoring distant prototypes. The generalized form of such an algorithm with different $l_p$-distance functions and the corresponding cluster location estimates was depicted in \cite{hamalainen2017comparison}.

\begin{algorithm}[t]
\caption{K-means$\parallel$}
\label{alg:K-meansparallel}
\begin{algorithmic}[1]
\REQUIRE Dataset $\X$, \#clusters $K$, and over-sampling factor $l$.
\ENSURE Set of prototypes $\C = \{\c_1,\c_2,...,\c_K\}$.
    \STATE $\C \gets$ select point $\mathbf{c}_1$ uniformly random from $\X$.
    \STATE $\psi \gets$ compute ${SSE(\C)}$.
	\FOR {$O ( \log ( \psi ))$ times} 
 	\STATE  $\C' \gets$ sample each point $\x \in \X $ independently with probability ${l \cdot\ {d(\x)}^2} / SSE(\C) $.
 	\STATE  $\C \gets \C \cup\ \C'$
    \ENDFOR
    \STATE For each $\x$ in $\C$ attach a weight defined as the number of points in $\X$ closer to $\x$ than any other point in $\C$.
    \STATE Do a weighted clustering of $\C$ into $K$ clusters.
 \end{algorithmic}
\end{algorithm}

The parallelized K-means++ method, called K-means$\parallel$, was proposed by Bahmani et al. \cite{Bah2012} (see Algorithm \ref{alg:K-meansparallel}). In the algorithm, one samples points from $\X$ in a slightly different fashion compared to K-means++. More precisely, the sampling probabilities are multiplied with the over-sampling factor $l$ and the sampling is done independently for each data point. The initial SSE for the first sampled point $\psi$ determines the number of sampling iterations. K-means$\parallel$ runs $O(log(\psi))$ sampling iterations. For each iteration, the expected number of points is $l$. Hence, after $O(log(\psi))$ iterations, the expected number of points added to $\C$ is $O(l\, log(\psi))$. Finally, weights representing the accumulation of data around the sampled points are set and the result of the weighted clustering then provides the $K$ initial prototypes. K-means++ can be used to cluster the weighted data (see Algorithm 1 in \cite{bachem2017distributed}). Selecting $r = 5$ instead of $O(log(\psi))$ rounds and setting the over-sampling factor to $2K$ were demonstrated to be sufficient in \cite{Bah2012}. Recently, Bachem et al. \cite{bachem2017distributed} proved theoretically that small $r$ instead of $O(log(\psi))$ iterations is sufficient in K-means$\parallel$. A modifcation of K-means$\parallel$ for initializing robust clustering was described and tested in \cite{HamKarRos2018}.

\begin{algorithm}[t]
\caption{SK-means$\parallel$}
\label{alg:SKmp}
\begin{algorithmic}[1]
\REQUIRE Subsets $\{\X_1,\X_2,...,\X_S\}$ , \#clusters $K$, and \#Lloyd's iterations $T_{init}$.
\ENSURE Set of prototypes $\C = \{\c_1,\c_2,...,\c_K\}$.
    \STATE $\C_i \gets$ for each subset $\X_{i}$ run K-means$\parallel$.
    \STATE $\C_i \gets$ for each subset $\X_{i}$ run $T_{init}$ Lloyd's iterations initialized with $\C_{i}$.
    \STATE Compute local SSE for each $\C_i$ in $\X_i$.
    \STATE $\C \gets$ select prototypes corresponding to smallest local SSE.
\end{algorithmic}
\end{algorithm}

\subsection{Random projection}

The background for RP \cite{Ach2003} comes from the Johnson-Lindenstrauss lemma \cite{johnson1984extensions}. The lemma states that points in a high dimensional space can be projected to a lower dimension space while approximately preserving the distances of the points, when the projection is done with a matrix whose elements are randomly generated. Hence, for an $N \times M$ dataset $\X$, let $\R \in M \times P$ be a random matrix. Then, the random projected data matrix $\widetilde{\X}$ is given by
$
\widetilde{\X}=  \frac{1}{\sqrt{P}} \X \R.
$
The random matrix $\R$ consists of independent random elements $(r_{ij})$ which can be drawn from one of the following probability distributions \cite{Ach2003}: 
$r_{ij} = +1$ with probability $1/2$, or $-1$ with probability $1/2$; or
$r_{ij} = +1$ with probability $1/6$, $0$ with probability $2/3$, or $-1$ with probability $1/6$.

\section{Reduced K-means$\parallel$ type initialization}\label{RedKmpp}

Next we introduce the novel initialization algorithms for K-means.

\subsection{SK-means$\parallel$}

The first new initialization method for K-means clustering, Subset K-means$\parallel$ (SK-means$\parallel$), is described in Algorithm \ref{alg:SKmp}. The method is based on $S$ randomly sampled non-disjoint subsets $\{\X_1,\X_2,...,\X_S\}$ from $\X$ of approximately equal size, such as $\X = \cup_{i=1}^S \X_i$. First, K-means$\parallel$ is applied in each subset, which gives the corresponding set of initial prototypes $\C_{i}$. Next, each initial prototype set $\C_{i}$ in ${\X}_{i}$ is refined with $T_{init}$ Lloyd's iterations. $T_{init}$ is assumed to be significantly smaller than the number of Lloyd's iterations needed for convergence. Then, SSE is computed locally for each $\C_i$ in $\X_i$. Differently from the earlier work \cite{HamKar2016}, this locally computed SSE is now used as the selection criteria for the initial prototypes instead of the global SSE. Computation of SSE for $\X_i$ in Step 3 is obviously much faster than to compute it for the whole $\X$. However, a drawback is that if the subsets are too small to characterize the whole data, the selection of the initial prototypes might fail. Therefore, $S$ should be selected such that the subsets are sufficiently large.

The convergence rate of K-means is fast and the most significant improvements in the clustering error are achieved during the first few iterations \cite{BotBen1995,Bro2014}. Therefore, for the initialization purposes, $T_{init}$ can restricted, e.g., to 5 iterations. Moreover, since the number of Lloyd's iterations needed for convergence might vary significantly (e.g., \cite{hamalainen2017comparison}), a restriction on the number of Lloyd's iterations helps in synchronization, when a parallel implementation of the SK-means$\parallel$ method is used.

The computational complexity of the K-means$\parallel$ method is of the order $\mathcal{O}(rlNM)$, where $r$ is the number of initialization rounds. Therefore, SK-means$\parallel$ also has the complexity of the order $\mathcal{O}(rlNM)$ in Step 1. In addition, SK-means$\parallel$ runs $T_{init}$ Lloyd's iterations with the complexity of $\mathcal{O}(T_{init}KNM)$, and computes local SSE with the complexity of $\mathcal{O}(KNM)$. Hence, the total complexity of SK-means$\parallel$ is of the order $\mathcal{O}(rlNM + T_{init}KNM)$. 

\begin{algorithm}[t]
\caption{SRPK-means$\parallel$}
\label{alg:SRPKmp}
\begin{algorithmic}[1]
\REQUIRE Subsets $\{\X_1,\X_2,...,\X_S\}$ , \#clusters $K$, \#Lloyd's iterations $T_{init}$, and random projection dimension $P$.
\ENSURE Set of prototypes $\C = \{\c_1,\c_2,...,\c_K\}$.
	\STATE $\R_{i} \gets$ for each subset $\X_{i}$ generate $M \times P$ random matrix.
	\STATE $\widetilde{\X}_{i} \gets$ for each subset $\X_{i}$ compute $\frac{1}{\sqrt{P}} \X_i \R_{i}$
    \STATE $\widetilde{\C}_{i} \gets$ for each $\widetilde{\X}_{i} $ run K-means$\parallel$.
    \STATE $\I_i \gets$ for each $\widetilde{\X}_{i}$ run $T_{init}$ Lloyd's iterations initialized with $\widetilde{\C}_{i}$.
    \STATE For each partitioning $\I_i$ compute prototypes $\C_i$ in original space $\X_i$.
    \STATE Compute local SSE for each $\C_i$ in $\X_i$.
    \STATE $\C \gets$ select prototypes corresponding to smallest local SSE.
\end{algorithmic}
\end{algorithm}

\subsection{SRPK-means$\parallel$}\label{SRPK}


The second novel proposal, Subset Random Projection K-means$\parallel$ (SRPK-means$\parallel$), adds RPs to SK-means$\parallel$. Since SK-means$\parallel$ mainly uses time in computing distances in Steps 1 and 2, it is reasonable to speed up the distance computation with RP. The RP based method is presented in Algorithm \ref{alg:SRPKmp}. Generally, SRPK-means$\parallel$ computes a set of candidate initial prototypes in a lower dimensional space and then evaluates these in the original space. Similarly to Algorithm \ref{alg:SKmp}, the best set of prototypes based on the local SSE are selected.

The proposal first computes a unique random matrix for each subset $\X_i$. Then, the $P$ dimensional random projected subset $\widetilde{\X}_{i}$ is computed in the each subset $\X_{i}$. Steps 3--4 are otherwise the same as the Steps 1--2 in Algorithm \ref{alg:SKmp}, but these steps are applied for the lower dimensional subsets $\{\widetilde{\X}_{1},\widetilde{\X}_{2},...,\widetilde{\X}_{S}\}$. Next, the labels $\I_i$ for partitioning each subset are used to compute $\C_i$ in the original space $\X_i$. Finally, the local SSEs are computed and the best set of prototypes are returned as the initial prototypes. Note that the last two steps in Algorithm \ref{alg:SRPKmp} are the same as Steps 3--4 in Algorithm \ref{alg:SKmp}. SRPK-means$\parallel$ computes projected data matrices, which require a complexity of $\mathcal{O}(PNM)$ (naive multiplication) \cite{Bou2010}. Execution of K-means$\parallel$ in the lower dimensional space requires $\mathcal{O}(rlNP)$, and $T_{init}$ Lloyd's iterations requires $\mathcal{O}(T_{init}KNP)$ operations. Step 6 requires $\mathcal{O}(KNM)$ operations, since it computes the local SSEs in the original space, so that the total computational complexity of the SRPK-means$\parallel$ method is $\mathcal{O}(PNM + rlNP + T_{init}KNP + KNM)$. Typically, applications of RP are based on the assumption $P << M$. Thus, when the dimension of data $M$ is increased, the contribution of the second and the third term of the total computational complexity start to diminish. Moreover, when both $M$ and $K$ are large compared to $P$, the last term dominates the overall computational complexity. Therefore, in terms of running time, SRPK-means$\parallel$ is especially suited for clustering large-scale data with very high dimensionality into a large number of clusters.

Fern and Brodley \cite{FerBro2003} noted that clustering with RP produces highly unstable and diverse clustering results. However, this can be exploited in clustering to find different candidate structures of data, which then can be combined into a single result \cite{FerBro2003}. The proposed initialization method in this paper uses a similar idea as it tries to find structures from multiple lower dimensional spaces that minimize the local SSE. In addition, selecting a result that gives the smallest local SSE excludes the bad structures, which could be caused by inappropriate $\R_i$ or $\C_i$.

\section{Parallel implementation of\newline \noindent\hangindent 0.65cm proposed algorithms}\label{ParImp}

Bahmani et al. \cite{Bah2012} implemented K-means$\parallel$ with the MapReduce programming model. It can also be implemented by the Single Program Multiple Data (SPMD) programming model with message passing. Then all the steps of the parallelized Algorithms \ref{alg:K-meansparallel}, \ref{alg:SKmp}, and \ref{alg:SRPKmp} are executed inside an SPMD block. Next, a parallel implementation of K-means$\parallel$ as depicted in Algorithm \ref{alg:K-meansparallel} is briefly described, by using Matlab Parallel Computing Toolbox (PCT), SPMD blocks, and message passing functions (see \cite{Sha2009} for a detailed description about PCT). First, data $\X$ is split into $Q$ subsets of approximately equal size and then the subsets are distributed to $Q$ workers. Step 1 picks a random point from a random worker and broadcasts this point to all other workers. In Step 2, each worker calculates distances and SSE for its local data. Next, points are aggregated by calling {\sl{gplus}}-function, after which the aggregation distributes this sum to other workers. In Steps 4--5, each worker samples points from its local data, the next points are aggregated to $\C'$ by calling {\sl{gop}}-function, and then $\C'$ is broadcasted to all workers. Again, distances and SSE are calculated similarly as in Step 2. Each worker in Step 6 assigns weights based on its local data, after which the weights are aggregated with {\sl{gop}}-function. Finally, Step 7 is computed sequentially.

Similarly to the parallel K-means$\parallel$ implementation, a parallel implementation of Algorithm \ref{alg:SKmp} with SMPD and message passing is described next. First, each subset $\X_i$ from $S$ subsets is split into $J$ approximately equal size subsets and then these subsets are distributed to $J \times S$ workers, e.g., subset $\X_i$ is distributed to workers $(i-1)J+1,...,(i-1)J+J$. In Steps 1--3, each subset of workers runs steps for subset $\X_i$ in parallel similarly as described in the previous paragraph. For parallel Lloyd's iterations, a similar strategy as proposed in \cite{Dhi1999} can be used in Step 2. Steps 1--3 require calling modified {\sl{gop}}-function and {\sl{gplus}}-function for the subset of workers; these functions were modified to support this requirement. Finally, prototypes corresponding to the smallest local SSE from the subset $i'$ allocated workers $(i'-1)J+1,...,(i'-1)J+J$ are returned as the initialization.

The parallel SRPK-means$\parallel$ in Algorithm \ref{alg:SRPKmp} can be implemented in a highly similar fashion to the parallel SK-means$\parallel$. More precisely, in Step 1, each worker $(i-1)J+1$, where $i \in \{1,2,...,S\}$, generates the random matrix $\R_i$ and broadcasts it to workers $(i-1)J+1,...,(i-1)J+J$. In Step 2, each worker computes random projected data for its local data. Steps 3--4 are otherwise computed similarly to the parallel SK-means$\parallel$ Steps 1--2, except these steps are executed for the projected subsets. In Step 5, the prototypes are computed in the original space in parallel. Finally, Steps 6--7 are the same as Steps 3--4 in Algorithm \ref{alg:SKmp}. The parallel implementations of the proposed methods and K-means$\parallel$ are available in \footnote{\url{https://github.com/jookriha/Scalable-K-means}}. 

\section{Empirical evaluation of\newline \noindent\hangindent 0.65cm proposed algorithms}

In this section, empirical comparison between K-means++, K-means$\parallel$, SK-means$\parallel$, and SRPK-means$\parallel$ is presented by using 21 datasets. In Section \ref{Sec:Results15datasets}, the results are given for 15 reference datasets. The performance of the methods was evaluated by analyzing SSE, the number of iterations needed for convergence, and the running time. Finally, In Section \ref{Sec:KSPHERES}, we analyze the final clustering accuracy for six novel synthetic datasets that highlight the effects of the curse of dimensionality in the K-means++ type initialization strategies. The simulated clustering problems have been formed with the novel generator described in Algorithm \ref{alg:gendata}. The MATLAB implementation of the algorithm is available in \footnote{\url{https://github.com/jookriha/M_Spheres_Dataset_Generator}}.



\subsection{Experiments with reference datasets}\label{Sec:Results15datasets}

In this section, the results are shown and analyzed for 15 publicly available reference datasets by considering separately the accuracy (Section \ref{Sec:SSE}), efficiency (Section \ref{Sec:eff}), and scalability (Section \ref{Sec:sca}) of the algorithms.

\subsubsection{Experimental setup}

 \begin{table}[t]
 \caption{Characteristics of datasets}\label{tab:datasets}
 \normalsize
 \centering
 \bgroup
 \def\arraystretch{1.2}%
 \begin{adjustbox}{max width=1\columnwidth}
 \begin{tabular}{crrr}
 \hline
 \text{Dataset} & $N$ & $M$ & ${K}$ \\ \hline
 HAR & 7 352 & 561 & 6 \\
 ISO & 7 797 & 617 & 26 \\
 LET & 20 000 & 16 & 26 \\
 GFE & 27 936 & 300 & 36 \\ 
 MNI & 70 000 & 784 & 10 \\ 
 BIR & 100 000 & 2 & 100 \\  
 BSM & 583 250 & 77 & 50* \\
 FCT & 581 012 & 54 & 7 \\
 SVH & 630 420 & 3 072 & 100* \\
 RCV & 781 265 & 1 000 & 350 \\
 USC & 2 458 285 & 68 & 100* \\
 KDD & 4 898 431 & 41 & 100* \\
 M8M & 8 100 000 & 784 & 265 \\
 TIN & 15 860 403 & 384 & 100* \\
 OXB & 16 334 970 & 128 & 100* \\
 \hline 
 \end{tabular}
 \end{adjustbox}
 \egroup
 \end{table}

\begin{table*}[ht!]
\caption{\#iterations} \label{tab:iters}
\huge
\centering
\bgroup
\def\arraystretch{1.2}%
\begin{adjustbox}{max width=1.0\linewidth}
\begin{tabular}{llllllllllllllll}
\hline
& HAR$^{**}$ & ISO$^{**}$ & LET$^{**}$ & GFE$^{**}$ & MNI$^{**}$ & BIR & BSM$^{**}$ & FCT$^{**}$ & SVH$^{**}$ & RCV$^{**}$ & USC & KDD & M8M$^{**}$ & TIN$^{**}$ & OXB$^{**}$ \\
\hline
K-means$++$ & 25.5 & 31.5 & 79 & 62 & 86 & 94 & 36 & 14 & - & -  & - & - & - & - & - \\
K-means$\parallel$ & 28.5 & 33.5 & 68.5 & 59 & 86 & 97 & 35 & 9.5 & 32.5 & 20 & 81 & 82 & 31 & 37.5 & 27.5 \\
SK-means$\parallel$ & 23.5 & \bf{25.5}$^{+*\dagger_{5}}$ & \bf{63}$^{+*\dagger_{5,10}}$ & 52$^{+}$ & 73 & \bf{86.5}~ & 25.5$^{+*\dagger_{5}}$ & \bf{1}$^{+*\dagger_{5-40}}$ & 28.5$^{\dagger_{5}}$ & 19  & 86 & 72 & \bf{24}$^{*\dagger_{5-20}}$ & \bf{33}$^{*\dagger_{5-20}}$ & \bf{22}$^{*\dagger_{5-40}}$ \\
SRPK-means$\parallel$ $P = 5$ & 27 & 35 & 77 & 60 & 87 & - & 35 & 6$^{+*}$ & 35 & 20.5 & 93.5 & 96.5 & 33 & 40 & 28.5 \\
SRPK-means$\parallel$ $P = 10$ & \bf{19}$^{+*}$ & 30 & 76.5 & 55.5 & 72.5 & - & 30.5 & 5$^{+*}$ & 32 & 20 & 83.5 & 78 & 31.5 & 38.5 & 28 \\
SRPK-means$\parallel$ $P = 20$ & 20$^{+*}$ & 30 & - & 53.5 & 83 & - & 26.5$^{+*\dagger_{5}}$ & 4$^{+*}$ & 30 & \bf{17}$^{*\dagger_{5}}$  & \bf{77} & \bf{69} & 30 & 39 & 27.5 \\
SRPK-means$\parallel$ $P = 40$ ~~~& \bf{19}$^{+*}$ & 28$^{\dagger_{5}}$ & - & \bf{50}$^{+}$ & \bf{66} & - & \bf{24}$^{+*\dagger_{5}}$ & 4$^{+*\dagger_{5,10}}$ & \bf{27}$^{\dagger_{5}}$ & \bf{17}$^{*\dagger_{5,10}}$  & 98.5 & 70 & 28.5$^{\dagger_{5}}$ & 35.5 & 28.5 \\
\hline
\end{tabular}
\end{adjustbox}
\egroup
\end{table*}

\begin{table}[ht!]
\caption{Running time for the initialization in seconds} \label{tab:initTime}
\huge
\centering
\bgroup
\def\arraystretch{1.2}%
\begin{adjustbox}{max width=1.0\columnwidth}
\begin{tabular}{lccccccc}
\hline
                                & KDD       & USC       & OXB       & TIN       & M8M       & RCV       & SVH \\\hline
K-means$\parallel$              & \bf{5.0}  &\bf{3.0}   & 26.0      & 52.1      & 98.5      & 14.1      & 13.7 \\
SK-means$\parallel$             & 8.7       & 5.0       & 39.1      & 65.8      & 145.5     & 20.6      & 17.3 \\
SRPK-means$\parallel$ $P = 5$   & 7.9       & 3.9       & \bf{23.7} & \bf{24.9} & \bf{32.2} & \bf{4.8}  & 3.4 \\
SRPK-means$\parallel$ $P = 40$  & 10.2      & 5.6       & 27.4      & 28.5      & 37.9      & 5.6       & \bf{3.3}\\
\hline
\end{tabular}
\end{adjustbox}
\egroup
\end{table}

Basic information about the datasets is shown in Table \ref{tab:datasets}. 
The parallel implementations of the proposed methods and K-means$\parallel$ (omitting K-means++ readily tested in \cite{Bah2012}) were applied to the seven largest datasets and serial implementations were used otherwise. 
For the serial experiments, we used the following eight datasets: Human Activity Recognition Using Smartphones\footnote{\label{note1}\url{http://archive.ics.uci.edu/ml/index.php}} (HAR), ISOLET\footnotemark[1] (ISO), Letter Recognition\footnotemark[1] (LET), Grammatical Facial Expressions\footnotemark[1] (GFE), MNIST\footnote{\url{http://yann.lecun.com/exdb/mnist/}} (MNI), Birch3\footnote{\url{http://cs.joensuu.fi/sipu/datasets/}} (BIR), Buzz in Social Media\footnotemark[1] (BSM), and Covertype\footnotemark[1] (COV). For the parallel experiments, the following seven large high-dimensional datasets were used: Street View House Numbers\footnote{\url{http://ufldl.stanford.edu/housenumbers/}} (SVH), RCV1v2 collection of documents\footnote{\label{note2}\url{https://www.csie.ntu.edu.tw/~cjlin/libsvmtools/datasets}} (RCV), US Census Data 1990\footnotemark[1] (USC), KDD Cup 1999 Data\footnotemark[1] (KDD), MNIST8M \footnotemark[5] (M8M), Tiny Images\footnote{\url{http://horatio.cs.nyu.edu/mit/tiny/data/}} (TIN), and Oxford Buildings\footnote{\url{http://www.robots.ox.ac.uk/~vgg/data/oxbuildings/}} (OXB). The BIR dataset \cite{franti2018} was selected to test SK-means$\parallel$ for low dimensional data. With the OXB dataset, we utilized the transformed dataset with 128-dimensional SIFT descriptors extracted from the original dataset. For the TIN dataset, we sampled a 20 percent subset from the Gist binary file (\path{tinygist80million.bin}), where {79302017} images are characterized with 384 dimensional Gist descriptors. The highest dimensional dataset was SVH, where we combined the training, testing, and validation subsets into a single dataset. We excluded the attack type feature (class label) from the KDD dataset, used the Twitter data for the BSM dataset, and restricted to the training dataset of the HAR dataset. For the RCV dataset, we used the full industries test set (350 categories) and selected 1000 out {47236} features with the same procedure as in \cite{de2009k}. For the M8M and the RCV datasets, we used the scaled datasets given in\footnotemark[5], all other datasets were min-max scaled into $[-1,1]$.

We fixed $S = 8$ in the experiments. This means that each dataset was randomly divided into 8 subsets, which were roughly of equal size. The parallel experiments were run in Matlab R2014a environment. Otherwise, Matlab 2018a environment was used. The parallel algorithms were implemented with Matlab Parallel Computing Toolbox with the SPMD blocks and message passing functions as discussed in Section \ref{ParImp}. The parallel experiments were run in a computer cluster utilizing Sandy Bridge nodes with 16 cores and 256 GB memory. A parallel pool of 32 workers was used in the experiments; therefore, 4 workers were allocated for each subset. In the parallel experiments, each worker had a $\frac{1}{4}$ random disjoint partition of the subset on the local workspace.

For all datasets we used the following settings: $i)$ for K-means$\parallel$: $l = 2K$ and $r = 5$; $ii)$ for SK-means$\parallel$ and SRPK-means$\parallel$: $T_{init} = 5$ and $S = 8$; $iii)$ and for SRPK-means$\parallel$: $P \in \{5,10,20,40\}$ and $R$ with $r_{ij} = \pm1$. Note that occurance of an empty cluster for SRPK-means$\parallel$ is possible in rare cases when all $S$ subsets produce an empty cluster. (For instance, empty clusters appeared seven times in all the experiments with the synthetic datasets as reported in Section \ref{Sec:KSPHERES}). In these cases, we repeated the whole clustering initialization from the start. After initialization, the Lloyd's algorithm was executed until the number of new assignments between the consecutive iterations was below or equal to the threshold. For the five largest datasets (SVH, RCV, OXB, M8M and TIN) we set this threshold to $1 \%$ of $N$ and otherwise to zero. In the parallel experiments, runs were repeated 10 times for each setting. In the serial experiments, runs were repeated 100 times for each setting. Values for the number of clusters, $K$, are given in the last column of Table \ref{tab:datasets}. Since the MNIST8M dataset is constructed artificially from the original MNIST dataset, we set $K$ for MNIST8M based on the optimal value for MNIST used by Gallego et al. \cite{gallego2018clustering}. Otherwise, the selection is either based on the known number of classes or fixed arbitrarily (indicated with * in Table \ref{tab:datasets}).

The quality of the clustering results between the methods was compared by using SSE. The SSE values were computed with formula \eqref{SSE} for the whole data. Finally, statistical comparison between the methods was performed with the nonparametric Kruskal-Wallis test \cite{kruskal1952use, SaaHamKar2017}, since in most of the cases the clustering errors were not normally distributed. The significance level was set to $0.05$.

\begin{table*}[p!]
\caption{Clustering accuracy using SSE. The statistically significant differences of the final SSE according to the Kruskal-Wallis test are indicated with ${**}$ in the first column. Symbols $+$, $*$, $\ddagger$, and $\dagger_{P'}$ indicate that the method has statistically significantly better SSE in a pairwise comparison with respect to K-means++ (K$++$), K-means$\parallel$ (K$\parallel$), SK-means$\parallel$ (SK$\parallel$), and SRPK-means$\parallel$ (SRPK$\parallel$) for $P = P'$, respectively. The coefficient under the name of the data in the first column is the data-specific multiplier which scales the SSE to the true level.} \label{tab:SSE}
\huge
\centering
\bgroup
\def\arraystretch{1.1}%
\begin{adjustbox}{max width=\textwidth}
\begin{tabular}{|c|c|c|c|c|c|c|c|c|c|c|c|c|c|c|c|c|c|c|c|c|c|}
\hline
& & \multicolumn{7}{c|}{Initialization} & \multicolumn{7}{c|}{Final} \\  \cline{3-16}
 \multirow{ 2}{*}{}& \multirow{ 2}{*}{} &  \multicolumn{1}{c|}{K$++$}  &  \multicolumn{1}{c|}{K$\parallel$}    & \multicolumn{1}{c|}{SK$\parallel$} & \multicolumn{4}{c|}{SRPK$\parallel$} &  \multicolumn{1}{c|}{K$++$} &  \multicolumn{1}{c|}{K$\parallel$} &\multicolumn{1}{c|}{SK$\parallel$}& \multicolumn{4}{c|}{SRPK$\parallel$} \\
data  & stats  & &  & &  $P = 5$ & $P = 10$ & $P = 20$ & $P = 40$ &  & & &  $P = 5$ & $P = 10$ & $P = 20$ & $P = 40$ \\ \hline
HAR$^{**}$& median  &\cellcolor{gray!25}2.5881&\cellcolor{gray!25}1.5577&\bf{1.3958}&1.4474&1.4192 & 1.4074 & 1.4003 &\cellcolor{gray!25}1.3803&\cellcolor{gray!25}1.3803&\bf{1.3707}$^{+*}$&\bf{1.3707}$^{+*}$&\bf{1.3707}$^{+*}$&\bf{1.3707}$^{+*}$&\bf{1.3707}$^{+*}$  \\
$10^5$ & mad  &\cellcolor{gray!25}0.1498&\cellcolor{gray!25}0.0436&\bf{0.0066}&0.0250& 0.0175 & 0.0132 & 0.0116 &\cellcolor{gray!25}0.0203 &\cellcolor{gray!25}0.0242 & \bf{0.0057} & 0.0117 & 0.0087 & 0.0082 & 0.0064  \\
& max  &\cellcolor{gray!25}3.2295&\cellcolor{gray!25}1.7367&\bf{1.4316}&1.5269 & 1.4613 & 1.4504 & 1.4621 &\cellcolor{gray!25}1.5461 &\cellcolor{gray!25}1.5461 & 1.4126 & 1.4136 & 1.4110 & 1.4126 & \bf{1.4094}  \\
& min  &\cellcolor{gray!25}2.2533&\cellcolor{gray!25}1.4782&\bf{1.3785}&1.3971 & 1.3907 & 1.3859 & 1.3825&\cellcolor{gray!25}\bf{1.3707}&\cellcolor{gray!25}\bf{1.3707}&\bf{1.3707}&\bf{1.3707}&\bf{1.3707}&\bf{1.3707}&\bf{1.3707}  \\ \hline
ISO$^{**}$& median  &\cellcolor{gray!25}8.9267&\cellcolor{gray!25}5.5274&\bf{4.9617}& 5.7887 & 5.4478 & 5.198 & 5.0575 &\cellcolor{gray!25}4.7811&\cellcolor{gray!25}4.7795&4.7617$^{+*}$&4.7578$^{+*}$&4.7558$^{+*}$&\bf{4.7538}$^{+*}$&4.7575$^{+*}$ \\
$10^5$ & mad  &\cellcolor{gray!25}0.1797&\cellcolor{gray!25}0.0575&\bf{0.0271}& 0.0927 & 0.0723 & 0.0486 & 0.0411 &\cellcolor{gray!25}0.0287 &\cellcolor{gray!25}0.0246 & \bf{0.0202} & 0.0277 & 0.0306 & 0.0265 & 0.0267  \\
& max  &\cellcolor{gray!25}9.4413&\cellcolor{gray!25}5.7650&\bf{5.0490}& 6.0687 & 5.6078 & 5.2900 & 5.1494 &\cellcolor{gray!25}4.9228 &\cellcolor{gray!25}4.8663 & 4.8529 & 4.8469 & 4.8824 & \bf{4.8407} & 4.8787  \\
& min  &\cellcolor{gray!25}8.4326&\cellcolor{gray!25}5.3859&\bf{4.8820}& 5.6207 & 5.2543 & 5.0629 & 4.9383 &\cellcolor{gray!25}4.7142 &\cellcolor{gray!25}4.7085 & 4.7087 & 4.7145 & 4.7099 & \bf{4.7084} & 4.7180  \\ \hline
LET$^{**}$& median  &\cellcolor{gray!25}1.7868&\cellcolor{gray!25}1.2356&\bf{1.1415}&1.3543&1.2339& -  & -  &\cellcolor{gray!25}1.1012&\cellcolor{gray!25}1.1014&\bf{1.0985}$^{*}$&1.0994&1.0989& -  & -    \\
$10^4$ & mad  &\cellcolor{gray!25}0.0517&\cellcolor{gray!25}0.0176&\bf{0.0070}&0.0372&0.0217& -  & -  &\cellcolor{gray!25}0.0062 &\cellcolor{gray!25}0.0060 & \bf{0.0051} & 0.0064 & 0.0065& -  & -    \\
& max  &\cellcolor{gray!25}2.1162&\cellcolor{gray!25}1.3133&\bf{1.1616}&1.4327&1.2991& -  & -  &\cellcolor{gray!25}1.1261 &\cellcolor{gray!25}1.1192 & 1.1218 & \bf{1.1175} & 1.1219& -  & -    \\
& min  &\cellcolor{gray!25}1.6656&\cellcolor{gray!25}1.1771&\bf{1.1175}&1.2617&1.1844& -  & -  &\cellcolor{gray!25}\bf{1.0872} &\cellcolor{gray!25}1.0873 & 1.0883 & 1.0876 & 1.0875& -  & -    \\ \hline
GFE$^{**}$& median  &\cellcolor{gray!25}3.0103&\cellcolor{gray!25}2.0860&\bf{1.9218}& 2.1637 & 2.0463 & 1.9813 & 1.9506 &\cellcolor{gray!25}1.8605&\cellcolor{gray!25}1.8550&1.8491$^{+}$&1.8398$^{+*\ddagger}$&\bf{1.8397}$^{+*\ddagger}$&1.8407$^{+*\ddagger}$&1.8420$^{+*\ddagger}$  \\
$10^5$ & mad  &\cellcolor{gray!25}0.0994&\cellcolor{gray!25}0.0231&\bf{0.0155}& 0.0415&0.0258&0.0727&0.0704&\cellcolor{gray!25}0.0151 &\cellcolor{gray!25}0.0146 & 0.0117 & 0.0129 & 0.0123 & 0.0119 & \bf{0.0113}  \\
& max  &\cellcolor{gray!25}3.3933&\cellcolor{gray!25}2.1737&\bf{1.9793}&2.2439&2.1042&2.0302&1.9990&\cellcolor{gray!25}1.9440 &\cellcolor{gray!25}1.9105 & 1.8999 & 1.8783 & \bf{1.8700} & 1.8757 & 1.8771  \\
& min  &\cellcolor{gray!25}2.7908&\cellcolor{gray!25}2.0247&\bf{1.8819}&2.0618 & 1.9802 & 1.9424 & 1.9039 &\cellcolor{gray!25}1.8252 &\cellcolor{gray!25}1.8197 & 1.8236 & \bf{1.8172} & 1.8211 & 1.8227 & 1.8195  \\ \hline
MNI$^{**}$& median  &\cellcolor{gray!25}1.9539&\cellcolor{gray!25}1.2495&\bf{1.1074}&1.2156&1.1752&1.1458 & 1.1279&\cellcolor{gray!25}1.1013&\cellcolor{gray!25}1.1013&\bf{1.0979}$^{+*}$&1.0980$^{+}$&\bf{1.0979}$^{+*}$&\bf{1.0979}$^{+}$&1.0980  \\
$10^7$ & mad  &\cellcolor{gray!25}0.0624&\cellcolor{gray!25}0.0152&\bf{0.0032}&0.0117&0.0093&0.0068&0.0048&\cellcolor{gray!25}0.0027&\cellcolor{gray!25}0.0024&\bf{0.0017}&0.0023&0.0018&0.0026&0.0027  \\
& max  &\cellcolor{gray!25}2.2474&\cellcolor{gray!25}1.3056&\bf{1.1171}& 1.2428 & 1.1968 & 1.1606 & 1.1390 &\cellcolor{gray!25}1.1146&\cellcolor{gray!25}1.1075&1.1052&\bf{1.1046}&1.1069&1.1105&1.1095  \\
& min  &\cellcolor{gray!25}1.8138&\cellcolor{gray!25}1.2131&\bf{1.1006}& 1.1885 & 1.1472 & 1.1296 & 1.1136 & \cellcolor{gray!25}\bf{1.0977}&\cellcolor{gray!25}\bf{1.0977}&\bf{1.0977}&\bf{1.0977}&\bf{1.0977}&\bf{1.0977}&\bf{1.0977}  \\ \hline
BIR$^{**}$& median  &\cellcolor{gray!25}3.0658&\cellcolor{gray!25}1.9912&\bf{1.8677}& -  & -  & -  & -  &\cellcolor{gray!25}1.8440 &\cellcolor{gray!25}1.8187&\bf{1.7781}$^{+*}$& -  & -  & -  & -    \\
$10^2$ & mad  &\cellcolor{gray!25}0.1375&\cellcolor{gray!25}0.0544&\bf{0.0269}& -  & -  & -  & -  &\cellcolor{gray!25}0.0469 &\cellcolor{gray!25}0.0451 & \bf{0.0261}& -  & -  & -  & -    \\
& max  &\cellcolor{gray!25}3.4694&\cellcolor{gray!25}2.3346&\bf{1.9533}& -  & -  & -  & -  &\cellcolor{gray!25}2.0238 &\cellcolor{gray!25}2.0943 & \bf{1.8973}& -  & -  & -  & -    \\
& min  &\cellcolor{gray!25}2.6292&\cellcolor{gray!25}1.8770&\bf{1.8074}& -  & -  & -  & -  &\cellcolor{gray!25}1.7438 &\cellcolor{gray!25}1.7409 & \bf{1.7248}& -  & -  & -  & -    \\ \hline
BSM$^{**}$& median  &\cellcolor{gray!25}1.9621&\cellcolor{gray!25}1.1802&\bf{1.0780}&1.4680& 1.1957 & 1.1038 & 1.1012 &\cellcolor{gray!25}1.1914$^{\dagger_{5}}$&\cellcolor{gray!25}1.1631$^{\dagger_{5}}$&\bf{1.0698}$^{+*\dagger_{5,10}}$&1.2124&1.1324$^{+\dagger_{5}}$&1.0800$^{+*\dagger_{5,10}}$&1.0903$^{+*\dagger_{5,10}}$  \\
$10^5$ & mad  &\cellcolor{gray!25}0.1943&\cellcolor{gray!25}0.0741&\bf{0.0376}& 0.1593 & 0.0779 & 0.0511 & 0.0564 &\cellcolor{gray!25}0.0818 &\cellcolor{gray!25}0.0684 & \bf{0.0374} & 0.1006 & 0.0741 & 0.0504 & 0.0577  \\
& max  &\cellcolor{gray!25}2.5054&\cellcolor{gray!25}1.4593&\bf{1.1776}&1.8995&1.4179&1.2237&1.2130&\cellcolor{gray!25}1.4205 &\cellcolor{gray!25}1.4553 & \bf{1.1684} & 1.5541 & 1.2875 & 1.1813 & 1.2063  \\
& min  &\cellcolor{gray!25}1.5003&\cellcolor{gray!25}0.9999&0.9908&1.1971&0.9996&\bf{0.9779}&0.9780&\cellcolor{gray!25}0.9929 &\cellcolor{gray!25}\bf{0.9616} & 0.9739 & 1.0283 & 0.9668 & 0.9706 & 0.9664  \\ \hline
FCT$^{**}$& median  &\cellcolor{gray!25}3.8766&\cellcolor{gray!25}2.2187&\bf{1.9273}& 2.2076 & 2.1004 & 2.0086 & 1.9773 &\cellcolor{gray!25}1.9801&\cellcolor{gray!25}2.0000&\bf{ 1.9132}$^{+*\dagger_{5-40}}$ &1.9781&1.9670$^{*}$ &1.9461$^{+*\dagger_{5}}$&1.9385$^{+*\dagger_{5,10}}$  \\
$10^6$ & mad  &\cellcolor{gray!25}0.3322&\cellcolor{gray!25}0.1013&\bf{0.0343}&0.0654&0.0581&0.0508&0.0380&\cellcolor{gray!25}0.0634 &\cellcolor{gray!25}0.0784 & \bf{0.0354} & 0.0542 & 0.0506 & 0.0466 & 0.0435  \\
& max  &\cellcolor{gray!25}5.5290&\cellcolor{gray!25}2.5274&\bf{2.0294}& 2.4008& 2.2494 & 2.1397 & 2.0698 & \cellcolor{gray!25}2.2457 &\cellcolor{gray!25}2.3487 & \bf{2.0293} & 2.1752 & 2.1434 & 2.0852 & 2.0601  \\
& min  &\cellcolor{gray!25}3.1329&\cellcolor{gray!25}1.9329&\bf{1.8645}&2.0808&2.0013&1.8657&1.8652&\cellcolor{gray!25}\bf{1.8644}&\cellcolor{gray!25}\bf{1.8644}&\bf{1.8644}&\bf{1.8644}&\bf{1.8644}&\bf{1.8644}&\bf{1.8644}  \\ \hline
SVH& median  &\cellcolor{gray!25} -  &\cellcolor{gray!25} 1.3559 & \bf{1.0855}& 1.1820 & 1.1464 & 1.1155 & 1.0992 &\cellcolor{gray!25} -  &\cellcolor{gray!25} \bf{1.0703} & 1.0704& 1.0704 & 1.0705 & 1.0706 & 1.0708   \\
$10^8$ & mad  &\cellcolor{gray!25} -  &\cellcolor{gray!25} 0.0636 & \bf{0.0014} & 0.0185 & 0.0149 & 0.0075 & 0.0042 &\cellcolor{gray!25} -  &\cellcolor{gray!25} \bf{0.0003} & 0.0004 & 0.0005 & 0.0005 & \bf{0.0003} & \bf{0.0003}   \\
& max  &\cellcolor{gray!25} -  &\cellcolor{gray!25} 1.5968 & \bf{1.0889} & 1.2279 & 1.1765 & 1.1290 & 1.1083 &\cellcolor{gray!25} -  &\cellcolor{gray!25} \bf{1.0711} & 1.0720& \bf{1.0711} & 1.0717 & 1.0714 & 1.0712   \\
& min  &\cellcolor{gray!25} -  &\cellcolor{gray!25} 1.3027 & \bf{1.0836} & 1.1639 & 1.1246 & 1.1082 & 1.0922 &\cellcolor{gray!25} -  &\cellcolor{gray!25} \bf{1.0696} & 1.0698& 1.0698 & 1.0700 & 1.0703 & 1.0703   \\ \hline
RCV$^{**}$& median &\cellcolor{gray!25} - &\cellcolor{gray!25}2.5506&\bf{2.1405}&2.5897&2.4864&2.3575&2.2313&\cellcolor{gray!25} - &\cellcolor{gray!25}2.0876&2.0913&2.0780&2.0757$^{*\ddagger}$&\bf{2.0702}$^{*\ddagger\dagger_{5}}$&2.0715$^{*\ddagger\dagger_{5}}$  \\
$10^5$& mad &\cellcolor{gray!25} - &\cellcolor{gray!25}0.0233&\bf{0.0022}&0.0138&0.0041&0.0045&0.0040&\cellcolor{gray!25} - &\cellcolor{gray!25}0.0027&0.0018&0.0019&\bf{0.0014}&0.0026&0.0022  \\
& max &\cellcolor{gray!25} - &\cellcolor{gray!25}2.5886&\bf{2.1427}&2.5979&2.4945&2.3652&2.2363&\cellcolor{gray!25} - &\cellcolor{gray!25}2.0922&2.0951&2.0823&2.0778&2.0767&\bf{2.0755}  \\
& min &\cellcolor{gray!25} - &\cellcolor{gray!25}2.4849&\bf{2.1345}&2.5647&2.4830&2.3523&2.2228&\cellcolor{gray!25} - &\cellcolor{gray!25}2.0812&2.0863&2.0764&2.0737&2.0688&\bf{2.0674}  \\ \hline
USC$^{**}$& median &\cellcolor{gray!25} - &\cellcolor{gray!25}1.7014&\bf{1.1936}&1.5530&1.3218&1.2284&1.1989&\cellcolor{gray!25} - &\cellcolor{gray!25}1.1903&1.1779&1.1736$^*$&\bf{1.1688}$^*$&1.1718$^*$&1.1709$^*$  \\
$10^7$ & mad &\cellcolor{gray!25} - &\cellcolor{gray!25}0.0394&\bf{0.0036}&0.0338&0.0217&0.0079&0.0037&\cellcolor{gray!25} - &\cellcolor{gray!25}0.0070&0.0057&0.0072&0.0091&0.0091&\bf{0.0049}  \\
& max &\cellcolor{gray!25} - &\cellcolor{gray!25}1.7671&\bf{1.2016}&1.6178&1.3542&1.2415&1.2038&\cellcolor{gray!25} - &\cellcolor{gray!25}1.2020&1.1908&\bf{1.1803}&1.1841&1.1869&1.1829  \\
& min &\cellcolor{gray!25} - &\cellcolor{gray!25}1.6110&\bf{1.1877}&1.4957&1.2799&1.2191&1.1934&\cellcolor{gray!25} - &\cellcolor{gray!25}1.1781&1.1712&1.1595&\bf{1.1566}&1.1602&1.1667  \\ \hline
KDD$^{**}$& median &\cellcolor{gray!25} -  &\cellcolor{gray!25} 30.5335 & 2.5466 & 3.1781 & 2.6651 & 2.5817 & \bf{2.5162}  &\cellcolor{gray!25} - &\cellcolor{gray!25} 2.5218 & 2.4726& 2.4853 & 2.4582$^{*}$ & 2.4755 & \bf{2.4529}$^{*}$   \\
$10^5$ & mad   &\cellcolor{gray!25} - &\cellcolor{gray!25} 7.3666 & \bf{0.0337} & 0.1019 & 0.0483 & 0.0562 & 0.0440 &\cellcolor{gray!25} - &\cellcolor{gray!25} 0.0372 & 0.0419 & 0.0698 & 0.0413 & 0.0464 & \bf{0.0316}   \\
& max   &\cellcolor{gray!25} - &\cellcolor{gray!25} 58.0452 & \bf{2.5962} & 3.3024 & 2.7083 & 2.6275 & 2.6367 &\cellcolor{gray!25} - &\cellcolor{gray!25} 2.6288 & 2.5432& 2.6241 & \bf{2.5223} & 2.5640 & 2.5406   \\
& min   &\cellcolor{gray!25} -&\cellcolor{gray!25} 22.2667 & 2.4803& 2.9935 & 2.5447 & \bf{2.4483} & 2.4878 &\cellcolor{gray!25} - &\cellcolor{gray!25} 2.4614 & 2.4084& 2.3961 & \bf{2.3927} & 2.4032 & 2.4330   \\ \hline
M8M$^{**}$& median &\cellcolor{gray!25} - &\cellcolor{gray!25}2.6631&\bf{2.2390}&2.9313&2.6415&2.4185&2.3153&\cellcolor{gray!25} - &\cellcolor{gray!25}2.2159&2.2154&\bf{2.2139}$^*$&\bf{2.2139}$^*$&2.2141$^*$&\bf{2.2139}$^*$  \\
$10^8$ & mad &\cellcolor{gray!25} - &\cellcolor{gray!25}0.0164&\bf{0.0009}&0.0286&0.0150&0.0071&0.0041&\cellcolor{gray!25} - &\cellcolor{gray!25}0.0012&0.0008&0.0010&0.0009&\bf{0.0005}&0.0009  \\
& max &\cellcolor{gray!25} - &\cellcolor{gray!25}2.6959&\bf{2.2412}&2.9835&2.6690&2.4299&2.3176&\cellcolor{gray!25} - &\cellcolor{gray!25}2.2203&2.2165&2.2162&2.2155&\bf{2.2145}&2.2157  \\
& min &\cellcolor{gray!25} - &\cellcolor{gray!25}2.6391&\bf{2.2376}&2.8782&2.6192&2.4091&2.3043&\cellcolor{gray!25} - &\cellcolor{gray!25}2.2143&2.2131&2.2128&\bf{2.2126}&2.2131&2.2132  \\ \hline
TIN& median   &\cellcolor{gray!25} -&\cellcolor{gray!25} 10.7568 & \bf{8.8782} & 9.7335 & 9.4194 & 9.2042 & 9.0595 &\cellcolor{gray!25} - &\cellcolor{gray!25} 8.8060 & 8.8065& \bf{8.8058} & 8.8075 & 8.8073 & 8.8073   \\
$10^7$ & mad   &\cellcolor{gray!25} - &\cellcolor{gray!25} 0.1498 & \bf{0.0057} & 0.1054 & 0.0879 & 0.0338 & 0.0139 &\cellcolor{gray!25} - &\cellcolor{gray!25} \bf{0.0012} & 0.0014 & 0.0016 & 0.0018 & 0.0016 & 0.0030   \\
& max   &\cellcolor{gray!25} - &\cellcolor{gray!25} 11.1649 & \bf{8.8812} & 9.9343 & 9.5627 & 9.2543 & 9.0841 &\cellcolor{gray!25} - &\cellcolor{gray!25} 8.8091 & \bf{8.8077} & 8.8091 & 8.8106 & 8.8093 & 8.8108   \\
& min   &\cellcolor{gray!25} - &\cellcolor{gray!25} 10.4721 & \bf{8.8623} & 9.5824 & 9.3339 & 9.1431 & 9.0474 &\cellcolor{gray!25} - &\cellcolor{gray!25} 8.8031 & 8.8029 & 8.8042 & 8.8047 & 8.8044 & \bf{8.8024}   \\ \hline
OXB$^{**}$& median   &\cellcolor{gray!25} - &\cellcolor{gray!25} 1.7678 & \bf{1.5375} & 1.6432 & 1.6249 & 1.6014 & 1.5829 &\cellcolor{gray!25} - &\cellcolor{gray!25} 1.5267 & 1.5268& \bf{1.5254}$^{*\ddagger}$ & 1.5256$^{*\ddagger}$ & 1.5255$^{*\ddagger}$ & 1.5255$^{*\ddagger}$   \\
$10^8$ & mad   &\cellcolor{gray!25} -&\cellcolor{gray!25} 0.0181 & \bf{0.0004} & 0.0055 & 0.0058 & 0.0028 & 0.0017 &\cellcolor{gray!25} - &\cellcolor{gray!25} 0.0006 & \bf{0.0002} & 0.0003 & 0.0005 & 0.0005 & 0.0004   \\
& max   &\cellcolor{gray!25} - &\cellcolor{gray!25} 1.8224 & \bf{1.5384} & 1.6575 & 1.6315 & 1.6082 & 1.5850 &\cellcolor{gray!25} - &\cellcolor{gray!25} 1.5286 & 1.5271& \bf{1.5258} & 1.5266 & 1.5262 & 1.5261   \\
& min   &\cellcolor{gray!25} - &\cellcolor{gray!25} 1.7506 & \bf{1.5367} & 1.6393 & 1.6162 & 1.5995 & 1.5797 &\cellcolor{gray!25} - &\cellcolor{gray!25} 1.5258 & 1.5259& 1.5251 & \bf{1.5250} & \bf{1.5250} & \bf{1.5250}   \\ \hline
\end{tabular}
\end{adjustbox}
\egroup
\end{table*}

\subsubsection{Results for clustering accuracy}\label{Sec:SSE}

\begin{figure*}[ht!]
\captionsetup[subfigure]{}
  \centering
  \begin{subfigure}{0.32\textwidth}
    \includegraphics[width=\textwidth]{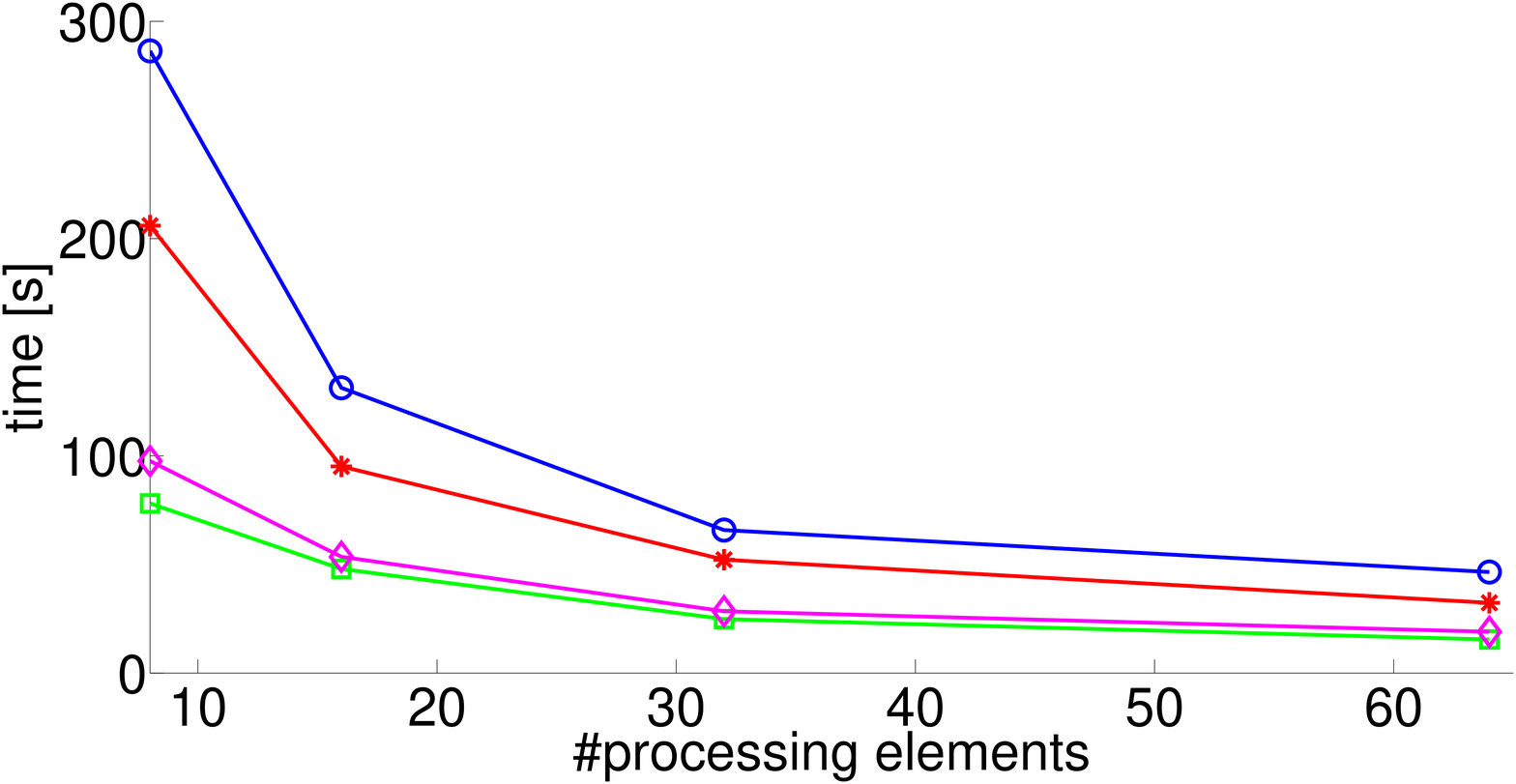}
    \caption{}
    \label{fig:amdahl}
  \end{subfigure} 
  ~
  \centering
  \begin{subfigure}{0.32\textwidth}
    \includegraphics[width=\textwidth]{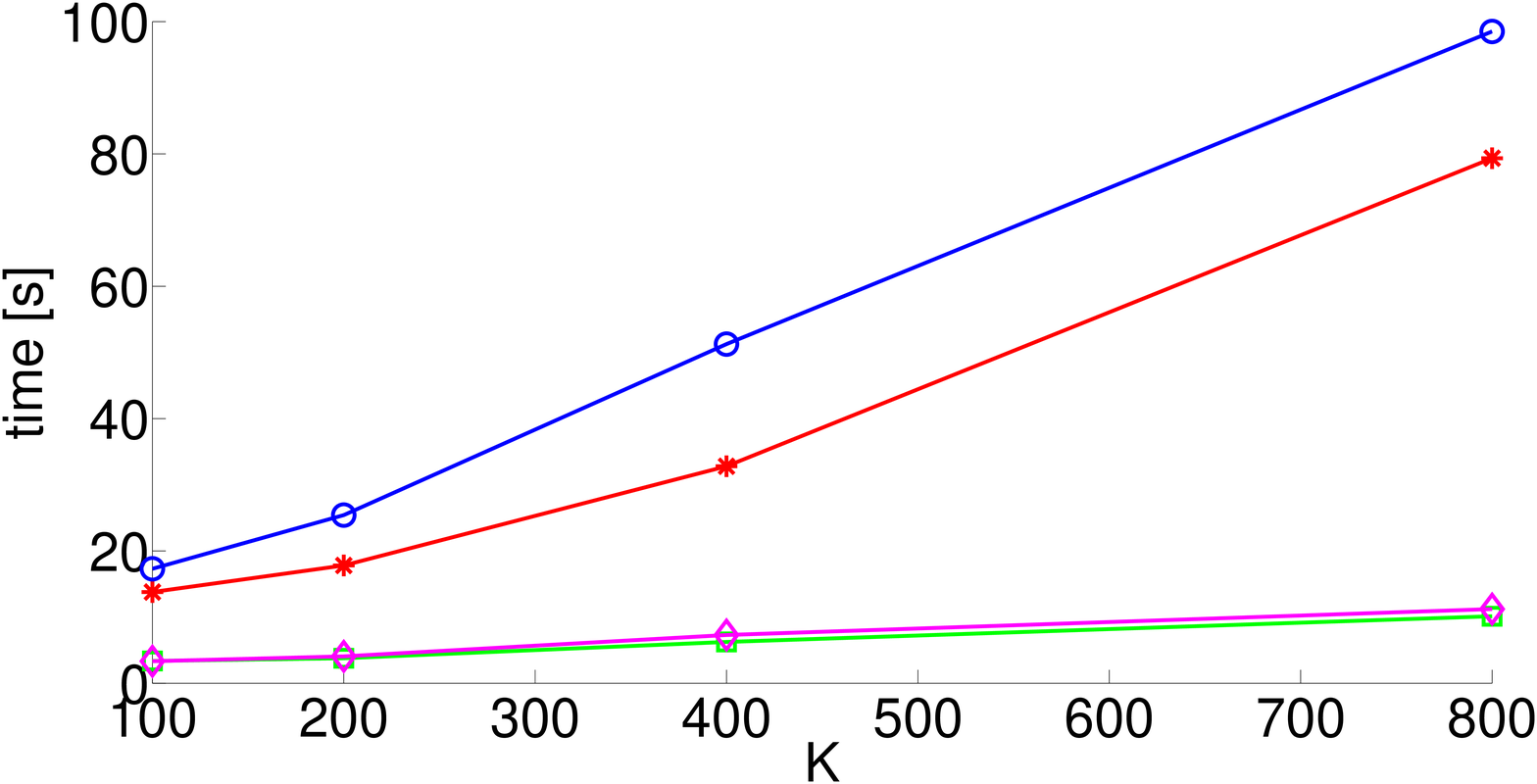}
    \caption{}
    \label{fig:runtime}
  \end{subfigure}
  ~
   \begin{subfigure}{0.32\textwidth}
    \includegraphics[width=\textwidth]{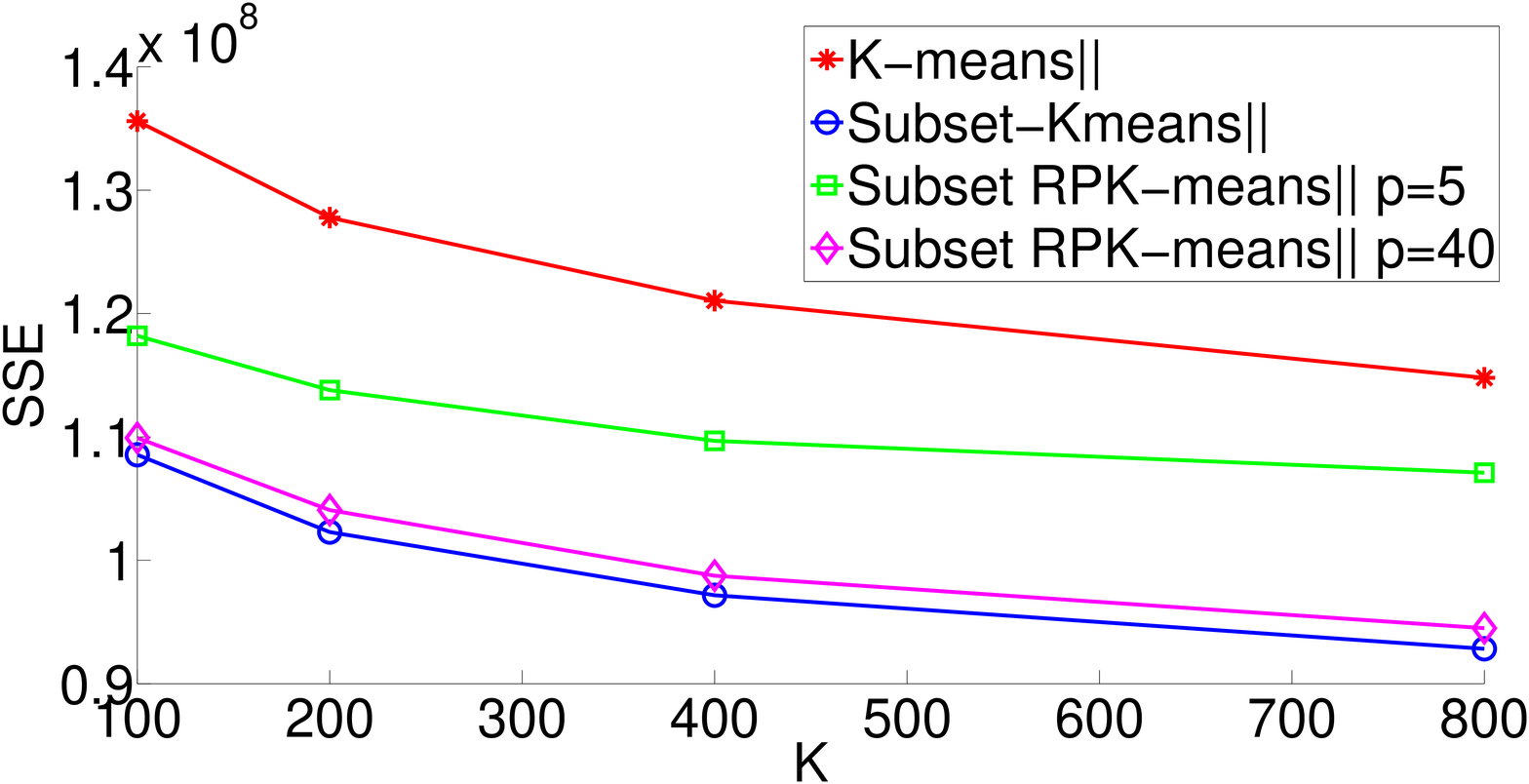}
    \caption{}
    \label{fig:sse}
  \end{subfigure}
  \caption{Scalability with respect to \#processing elements and the number of clusters $K$.}
  \label{fig:scalability}
\end{figure*}

SSE values after the initialization (initial SSE) and after the Lloyd's iterations (final SSE) are summarized in Table \ref{tab:SSE}. For the initial SSE, we did not include any results from the statistical testing because of differences in variances. Moreover, note that the assumption of equal variances of the final SSEs underlying the Kruskal-Wallis test, as tested with the Brown-Forsythe test, was only satisfied for SVH, RCV, USC, KDD, OXB, and M8M. For most datasets (HAR, ISO, LET, GFE, MNI, BIR, BSM, FCT, and TIN), this assumption was not satisfied.

Clearly, SK-means$\parallel$ and SRPK-means$\parallel$ outperform K-means$\parallel$ and K-means++ in terms of the initial SSE. SRPK-means$\parallel$ with $P = 40$ reaches almost the same initialization SSE level as SK-means$\parallel$. For the six largest datasets, SRPK-means$\parallel$ with $P = 20$ always had smaller max-value of SSE after initialization than the min-value of K-means$\parallel$. K-means++ has about two times larger initial SSE than SK-means$\parallel$ and SRPK-means$\parallel$ ($P = 40$). 

For all other datasets than SVH and TIN, the final clustering accuracy (SSE) was statistically significantly different between the four methods. Overall, in terms of the final clustering error, SRPK-means$\parallel$ achieved better final SSE than K-means$\parallel$ and K-means++. Moreover, one can note that for 11 out of 15 datasets, the random projection based initialization was better than the main baseline K-means$\parallel$. In many cases, SK-means$\parallel$ gives better final SSE than the baseline methods, but for the high-dimensional datasets, the results are equally good compared to the baseline methods. One can notice, based on the the statistical testing, that the final SSE is highly similar for K-means++ and K-means$\parallel$. Note that in Table \ref{tab:SSE} the min-values of all methods for the final SSE are equal for small number of clusters ($K \le 10$). This is probably due to the fact that the smaller number of possible partitions \cite{Ayramo2006} implies a smaller number of local minima compared to higher values of $K$.

\subsubsection{Results for running time and convergence}\label{Sec:eff}

Running time for the initialization (median of 10 runs) for the parallel experiments is shown in Table \ref{tab:initTime}. Running time for the initialization taken by K-means$\parallel$ is around $60\%-80\%$ of the running time of SK-means$\parallel$. SRPK-means$\parallel$ runs clearly faster than SK-means$\parallel$ for datasets with dimensionality more than 100, and for the four highest dimensional datasets, SRPK-means$\parallel$ runs clearly faster than K-means$\parallel$. Note that differences are small between $P = 5$ and $P = 40$ for SRPK-means$\parallel$.


The median number of Lloyd's iterations needed for convergence after the initialization phase are summarized in Table \ref{tab:iters}, where the statistically significant differences are denoted similarly as in Table \ref{tab:SSE}. The assumption of equal variances was satisfied for all datasets except for FCT. In general, SK-means$\parallel$ seems to  require smaller number of Lloyd's iterations than K-means++ and K-means$\parallel$, which directly translates to faster running time of the K-means search. Based on the statistical testing, SRPK-means$\parallel$ is better than or equal compared to the baseline methods in terms of the number of iterations. Therefore, SRPK-means$\parallel$ can also speed up the search phase of the K-means clustering method. Increasing the RP dimension from 5 to 40 further improved the speed of convergence for SRPK-means$\parallel$. Out of the parameter values used in the experiments, selecting $P = 40$ gives the best trade off between the running time and the clustering accuracy for SRPK-means$\parallel$. Furthermore, note that there is no statistical difference between K-means++ and K-means$\parallel$ with respect to the number of Lloyd's iterations.

\subsubsection{Results for scalability}\label{Sec:sca}

We conducted scalability tests for TIN and SVH to show how running time varies as a function of \#processing elements (Matlab workers) and to demonstrate the benefits of using SRPK-means$\parallel$ for a very high-dimensional dataset (SVH) when $K$ is increased. We concentrated on the running time of the initialization and the corresponding SSE. 
We performed scalability experiments in two parts: $1)$ Tests with TIN: \#processing elements was varied from 8 to 64 and $K = 100$ was fixed; $2)$ Tests with SVH: The number of clusters was varied as $K \in \{100,200,400,800\}$ and \#processing elements was fixed to 32. Otherwise, we used the same parameter settings as in the previous experiments. 

Median running time and SSE curves out of 10 runs are shown in Figure \ref{fig:scalability}. Results for the experiment $1$ are shown in Figure \ref{fig:amdahl}. In terms of Amdahl's law, K-means$\parallel$ and SK-means$\parallel$ perform equally well: running time is nearly halved when \#processing elements is doubled from 8 to 16 and from 16 to 32. In this perspective, performance of SRPK-means$\parallel$ is slightly worse than K-means$\parallel$ and SK-means$\parallel$. The results for the experiment $2$ are shown in Figure \ref{fig:runtime}--\ref{fig:sse}. Clearly, for very high-dimensional data, SRPK-means$\parallel$ runs much faster compared to K-means$\parallel$ and SK-means$\parallel$. As analyzed in Section \ref{SRPK}, the speedup for SRPK-means$\parallel$ is increased when $K$ is increased. A similar observation was made between K-means++ and RPK-means++ in \cite{chan2017efficient}. Furthermore, when $K = 800$, the speedup for SRPK-means$\parallel$ with respect to K-means$\parallel$ is 7--8 and with respect to SK-means$\parallel$ 9--10. Moreover, according to Figure \ref{fig:sse}, SRPK-means$\parallel$ (when $P = 40$) and SK-means$\parallel$ sustain their accuracy when $K$ is increased in a frame of K-means$\parallel$.

\begin{figure*}[ht]
\captionsetup[subfigure]{}
  \centering
  \begin{subfigure}{0.3\textwidth}
  \includegraphics[width=\textwidth]{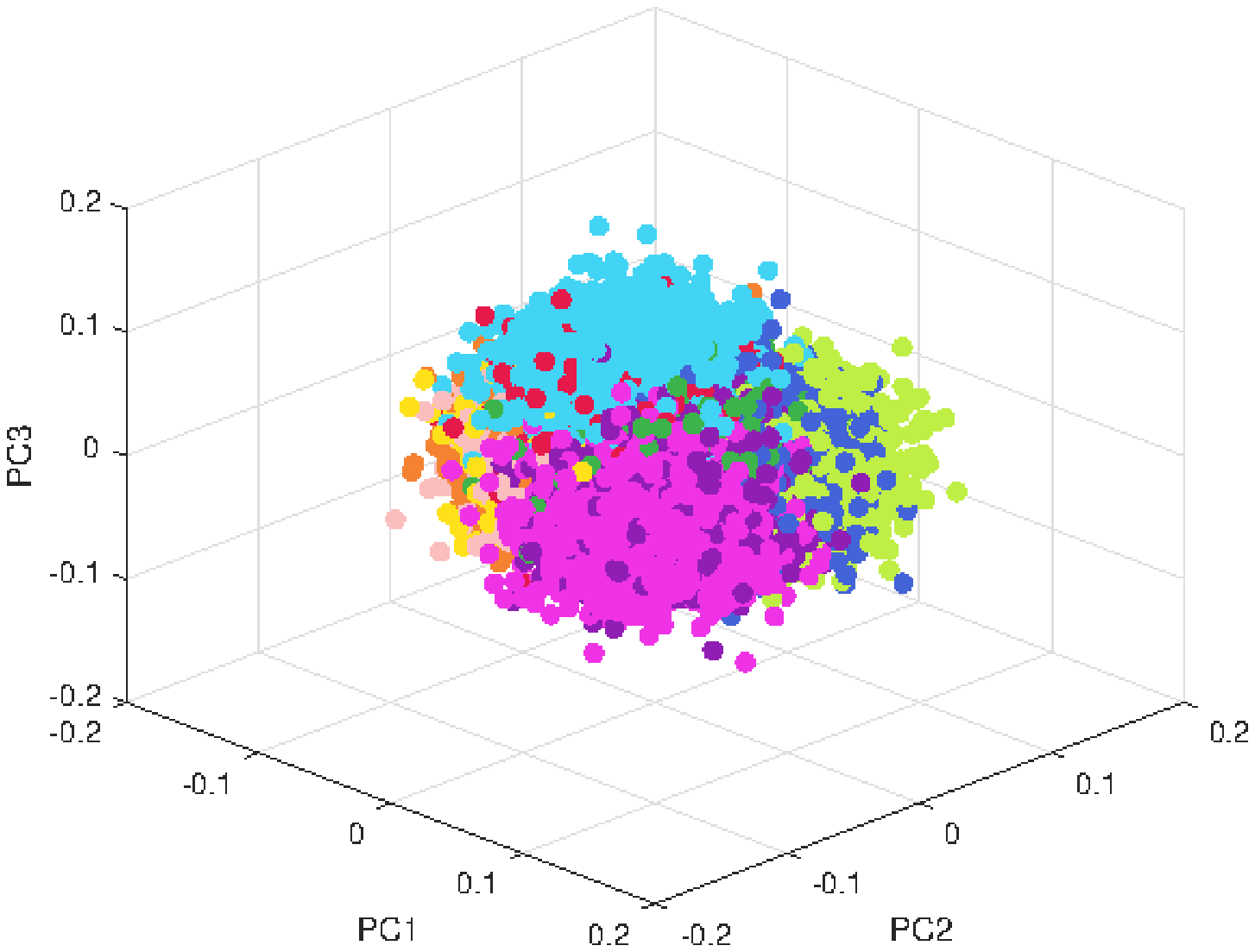}
    \caption{M-spheres-M1k-$d_c$0.05}
    \label{fig:PCA1}
  \end{subfigure} 
  ~
  \centering
  \begin{subfigure}{0.3\textwidth}
    \includegraphics[width=\textwidth]{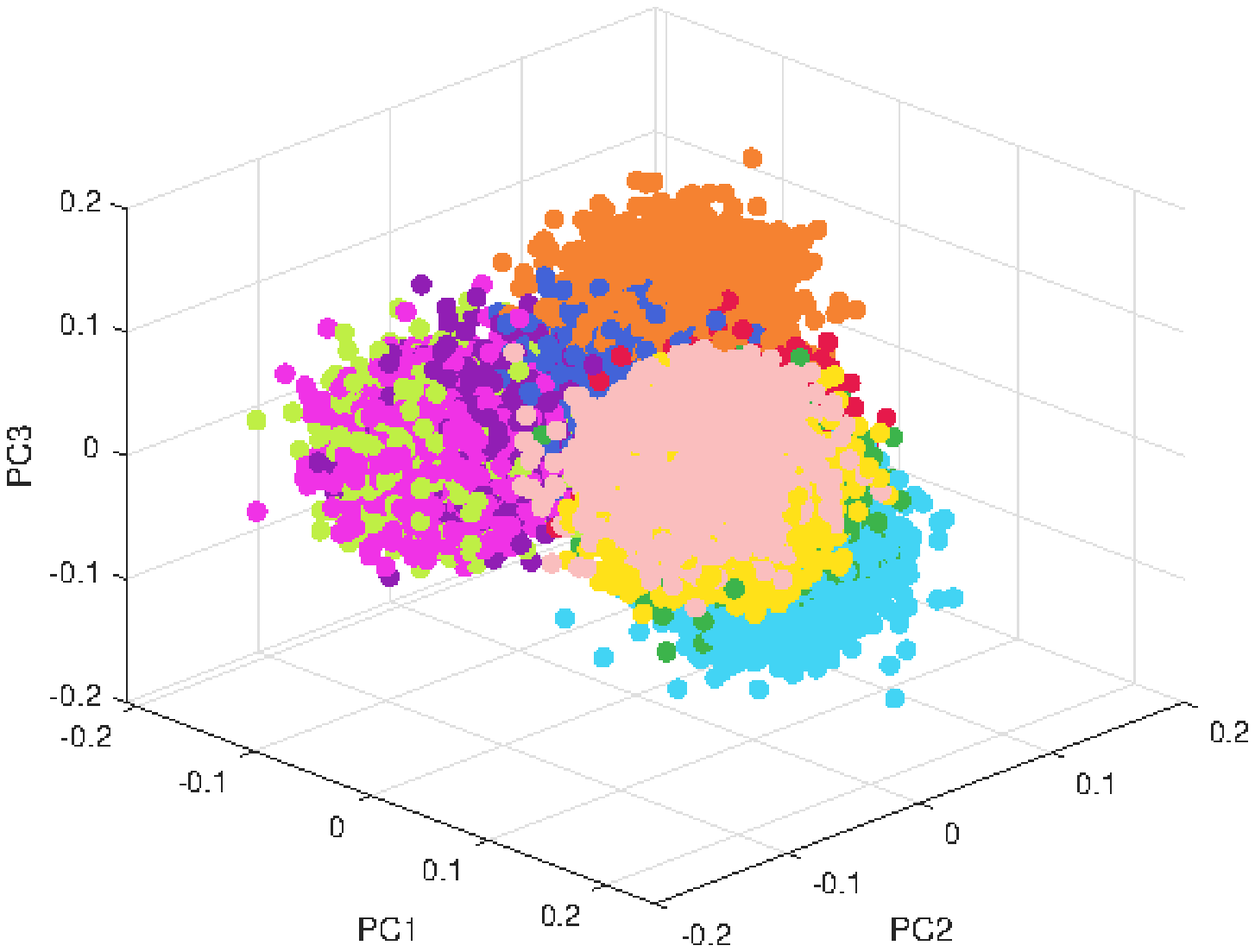}
    \caption{M-spheres-M1k-$d_c$0.1}
    \label{fig:PCA3}
  \end{subfigure}
  ~
  \centering
  \begin{subfigure}{0.3\textwidth}
    \includegraphics[width=\textwidth]{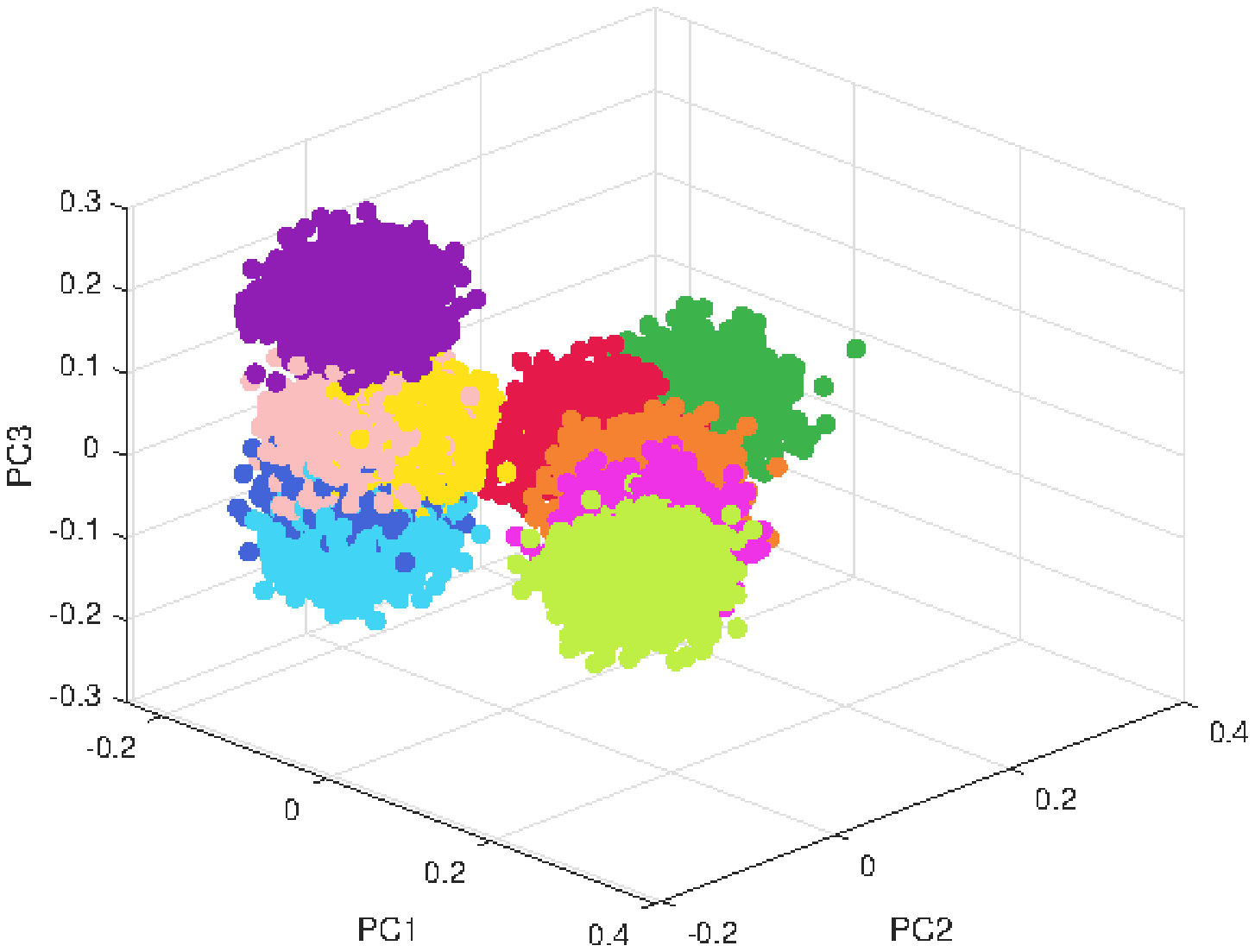}
    \caption{M-spheres-M1k-$d_c$0.2}
    \label{fig:PCA5}
  \end{subfigure} 
  
  \centering
  \begin{subfigure}{0.3\textwidth}
    \includegraphics[width=\textwidth]{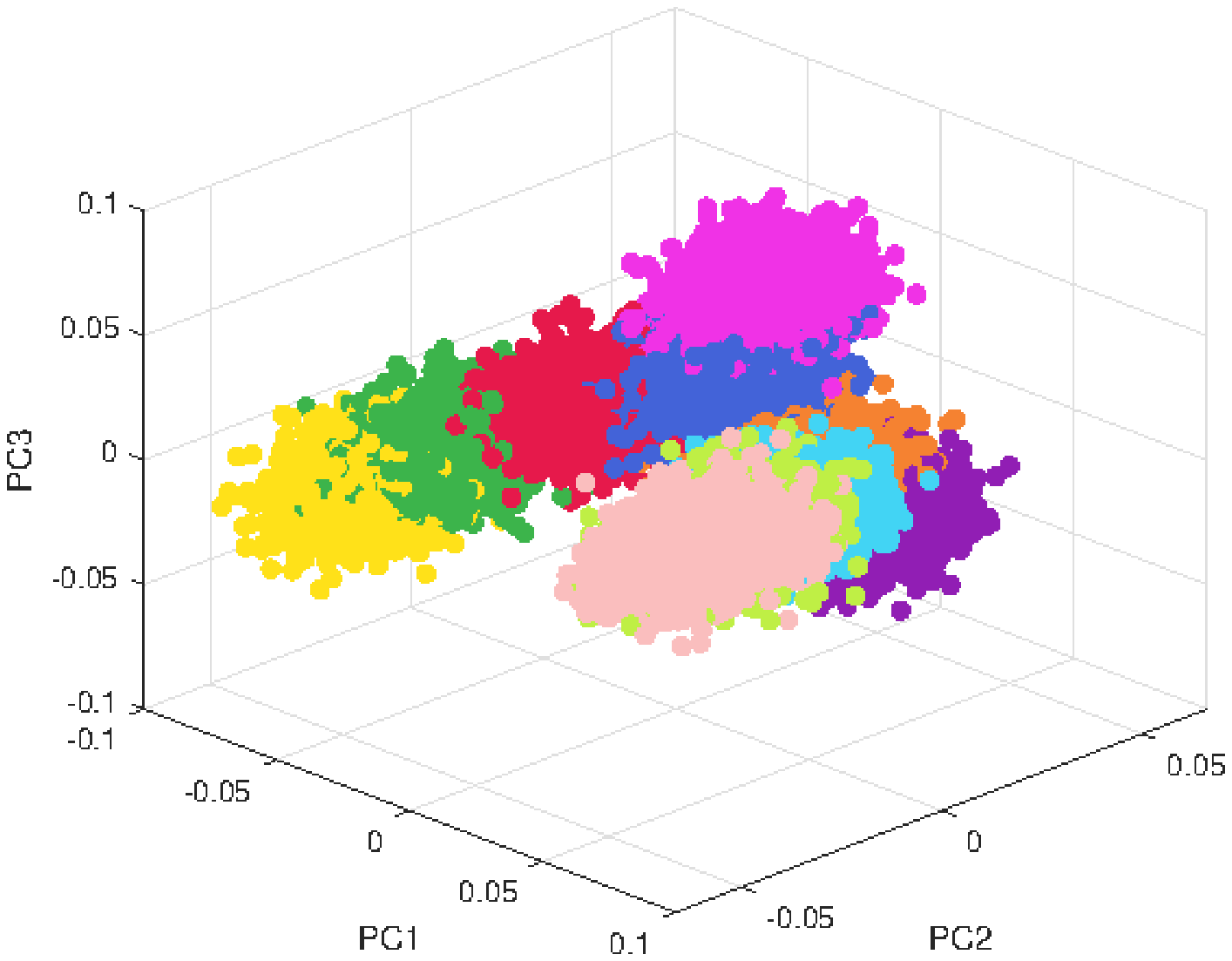}
    \caption{M-spheres-M10k-$d_c$0.05}
    \label{fig:PCA2}
  \end{subfigure}
  ~
  \centering
  \begin{subfigure}{0.3\textwidth}
    \includegraphics[width=\textwidth]{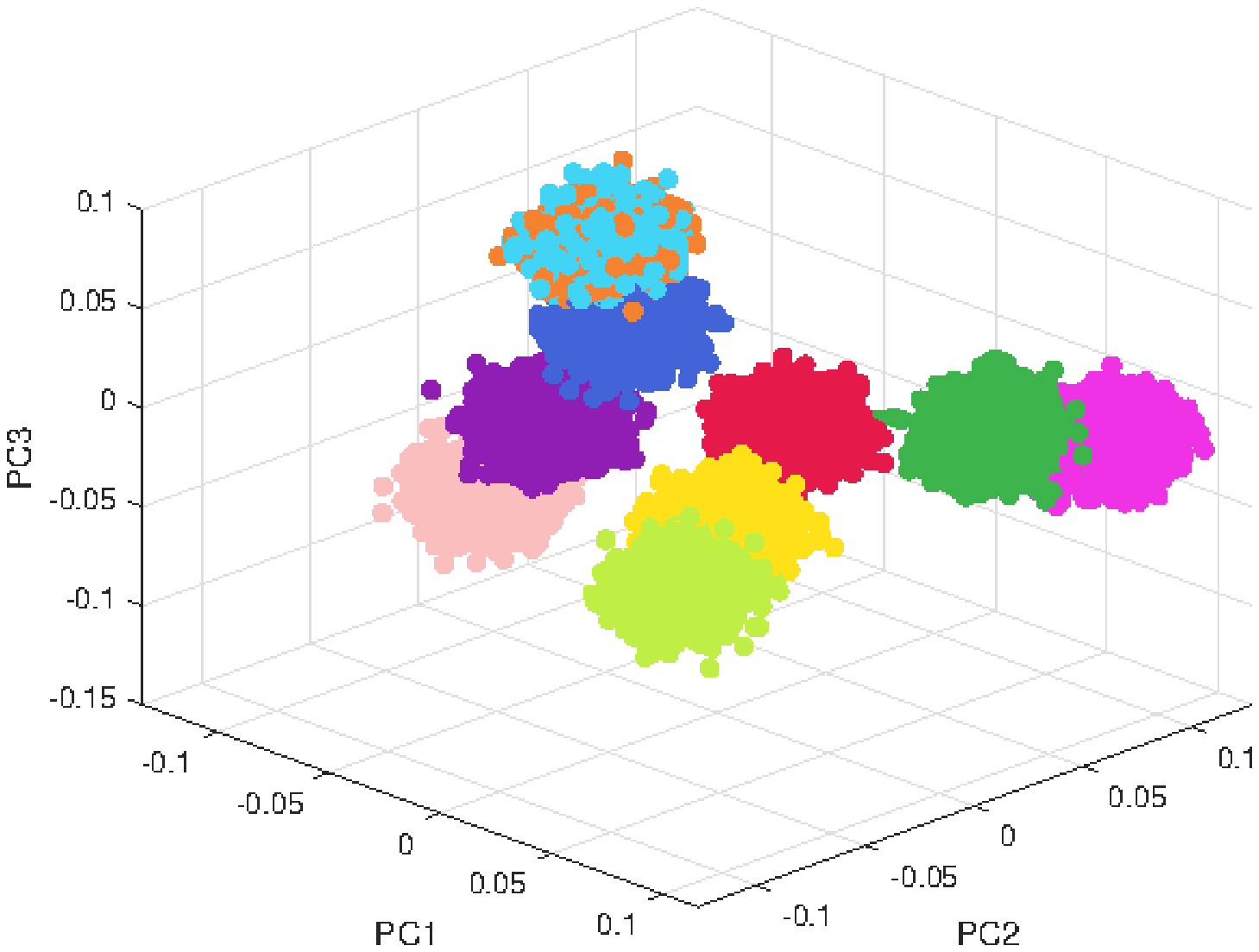}
    \caption{M-spheres-M10k-$d_c$0.1}
    \label{fig:PCA4}
  \end{subfigure} 
  ~
  \centering
  \begin{subfigure}{0.3\textwidth}
    \includegraphics[width=\textwidth]{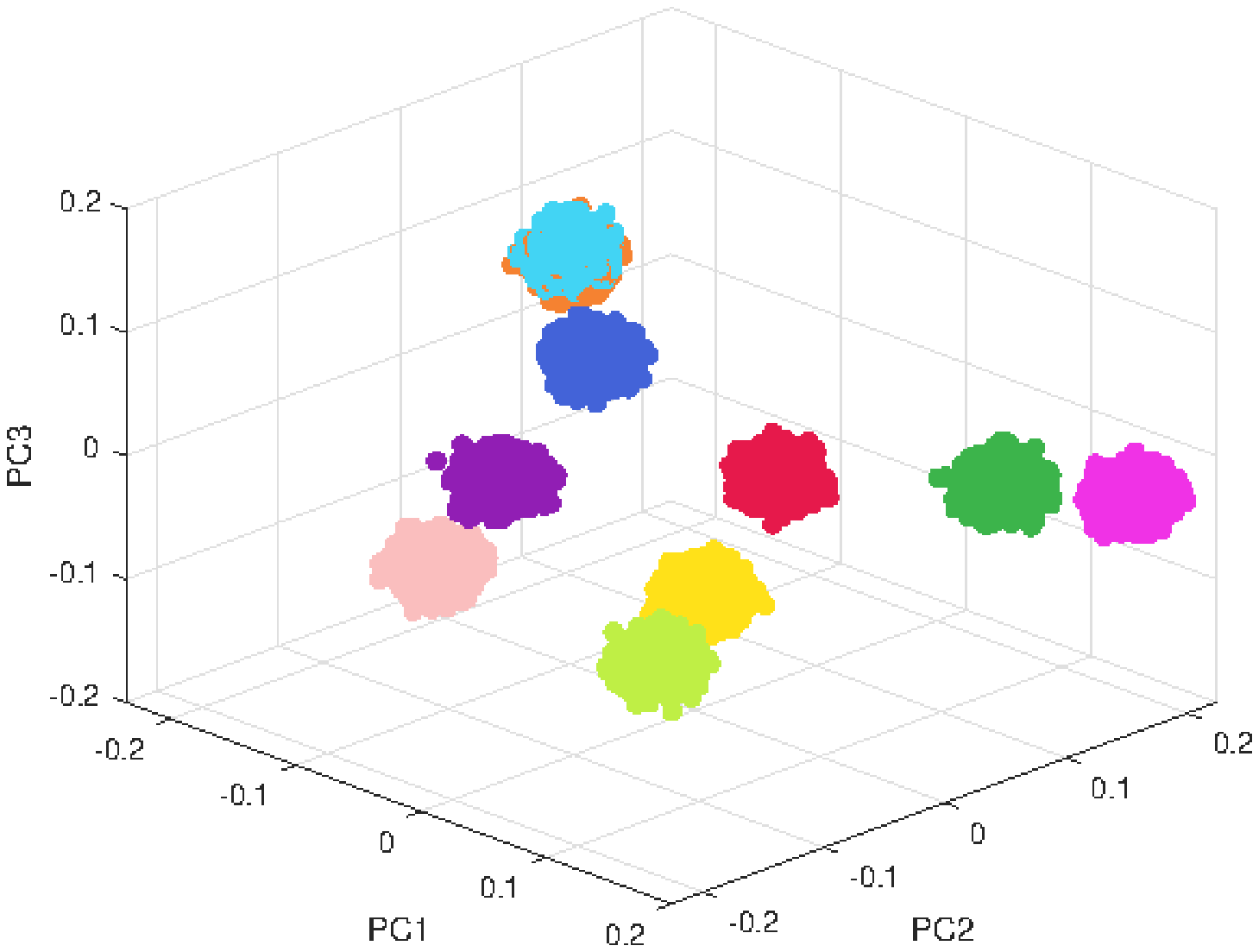}
    \caption{M-spheres-M10k-$d_c$0.2}
    \label{fig:PCA6}
  \end{subfigure} 
  \caption{Synthetic dataset projections on the three largest principal components.} 
  \label{fig:KSPHERES_PCA}
\end{figure*}

\begin{algorithm}[ht]
\caption{$M$-spheres dataset generator}
\label{alg:gendata}
\begin{algorithmic}[1]
\REQUIRE \#clusters $K$, \#dimensions $M$, \#points per cluster $N_K$, nearest center distance $d_c$, radius of M-sphere $d_r$.
\ENSURE Dataset $\X = \{\x_1,\x_2,...,\x_{N}\}$.
	\STATE $\C \gets$ \{(0,...,0)\}.
	\IF{$K > 1$} \STATE{
	        $\c_{2} \gets$ randsurfpoint($\c_{1}$,$d_c$).}
	        \STATE{$\C \gets$ $\C \cup\ \{\c_{2}\}$.}
	        \STATE{$k \gets 2$.}
    \ENDIF
	\IF{$K > 2$} 
	     \WHILE{$k < K$} 
	        \STATE{$i \gets$ rand($\{1,2,...,k\}$)}.
            \STATE{$\c_{cand} \gets$ randsurfpoint($\c_{i}$,$d_c$).}.
            \STATE{$i^* \gets \argmin_{j} \|\c_{j}-\c_{cand}\|$}.
            \IF{$i^* == i$} 
                \STATE{$\C \gets$ $\C \cup\ \{\c_{cand}\}$.}
                \STATE{$k \gets k + 1$.}
            \ENDIF
	     \ENDWHILE
     \ENDIF
     \STATE $k \gets 1$.
     \STATE $\X \gets \{\}$.
     \WHILE{$k \le K$}
     \STATE $n \gets 1$
        \WHILE{$n \le N_K$}
	        \STATE{$d_r^* \gets rand((0,d_r])$}
	        \STATE{$\x_{new} \gets$ randsurfpoint($\c_{k}$,$d_r^*$)}.
            \STATE{$\X \gets \X \cup\ \{\x_{new}\}$}.
            \STATE{$n \gets n + 1$}
        \ENDWHILE
     \STATE $k \gets k + 1$
     \ENDWHILE
\end{algorithmic}
\end{algorithm}

\begin{algorithm}[ht]
\caption{randsurfpoint($\c$,$d$)}
\label{alg:randsurfpoint}
\begin{algorithmic}[1]
\REQUIRE Sphere center $\c$, sphere radius $d$.
\ENSURE New point $\x^*$.
    \STATE $m \gets 1$.
    \STATE $S \gets 0$.
    \WHILE{$m \le M$}
        \STATE $x_m \gets \mathcal{N}(0,1)$.
        \STATE $S \gets S + x_m^2$.
        \STATE $m \gets m + 1$.
    \ENDWHILE
    \STATE{$\x^* \gets \c + d S^{-1/2} \x$}
\end{algorithmic}
\end{algorithm}

\begin{table}[ht]
\caption{Characteristics of the synthetic datasets}\label{tab:synthdatasets}
\normalsize
\centering
\bgroup
\def\arraystretch{1.2}%
\begin{adjustbox}{max width=1.0\columnwidth}
\begin{tabular}{lrrrrr}
\hline
\text{Dataset} & $N$ & $M$ & ${K}$ & $d_c$ & $d_r$  \\ \hline
M-spheres-M1k-$d_c$0.05 & 100 000 & 1 000 & 10 & 0.05 & 1.0 \\
M-spheres-M1k-$d_c$0.1 &  100 000 & 1 000 & 10 & 0.1 & 1.0 \\
M-spheres-M1k-$d_c$0.2 & 100 000 & 1 000 & 10 & 0.2 & 1.0 \\
M-spheres-M10k-$d_c$0.05 & 100 000 & 10 000 & 10 & 0.05 & 1.0 \\
M-spheres-M10k-$d_c$0.1 & 100 000 & 10 000 & 10 & 0.1 & 1.0 \\
M-spheres-M10k-$d_c$0.2 & 100 000 & 10 000 & 10 & 0.2 & 1.0 \\
\hline 
\end{tabular}
\end{adjustbox}
\egroup
\end{table}

\subsection{Experiments with high-dimensional\newline \noindent\hangindent 0.8cm synthetic datasets}\label{Sec:KSPHERES}

Finally, to strengthen the evaluation, we next summarize experiments with novel synthetic datasets, where symmetric, spherical clusters are hidden in a very high-dimensional space. Comparison of K-means initialization methods for datasets with assured spherical shape is clearly relevant because K-means restores such geometries during the clustering process \cite{Bah2012,hamalainen2017comparison,franti2018}. Note that using SSE for high-dimensional data can be ambiguous \cite{aggarwal2001surprising}. As we will next demonstrate, the SSE error difference of good and bad clustering results in a high-dimensional space can be surprisingly small. To show this, we also analyze the final clustering accuracy with normalized mutual information (NMI) \cite{strehl2002relationship} with the actual entropies. Note that it would have been uninformative to use NMI as a quality measure for datasets in Section \ref{Sec:Results15datasets} because there we had no information on the cluster geometry.

\begin{figure*}[ht]
\captionsetup[subfigure]{}
  \centering
  \begin{subfigure}{0.48\textwidth}
    \includegraphics[width=\textwidth]{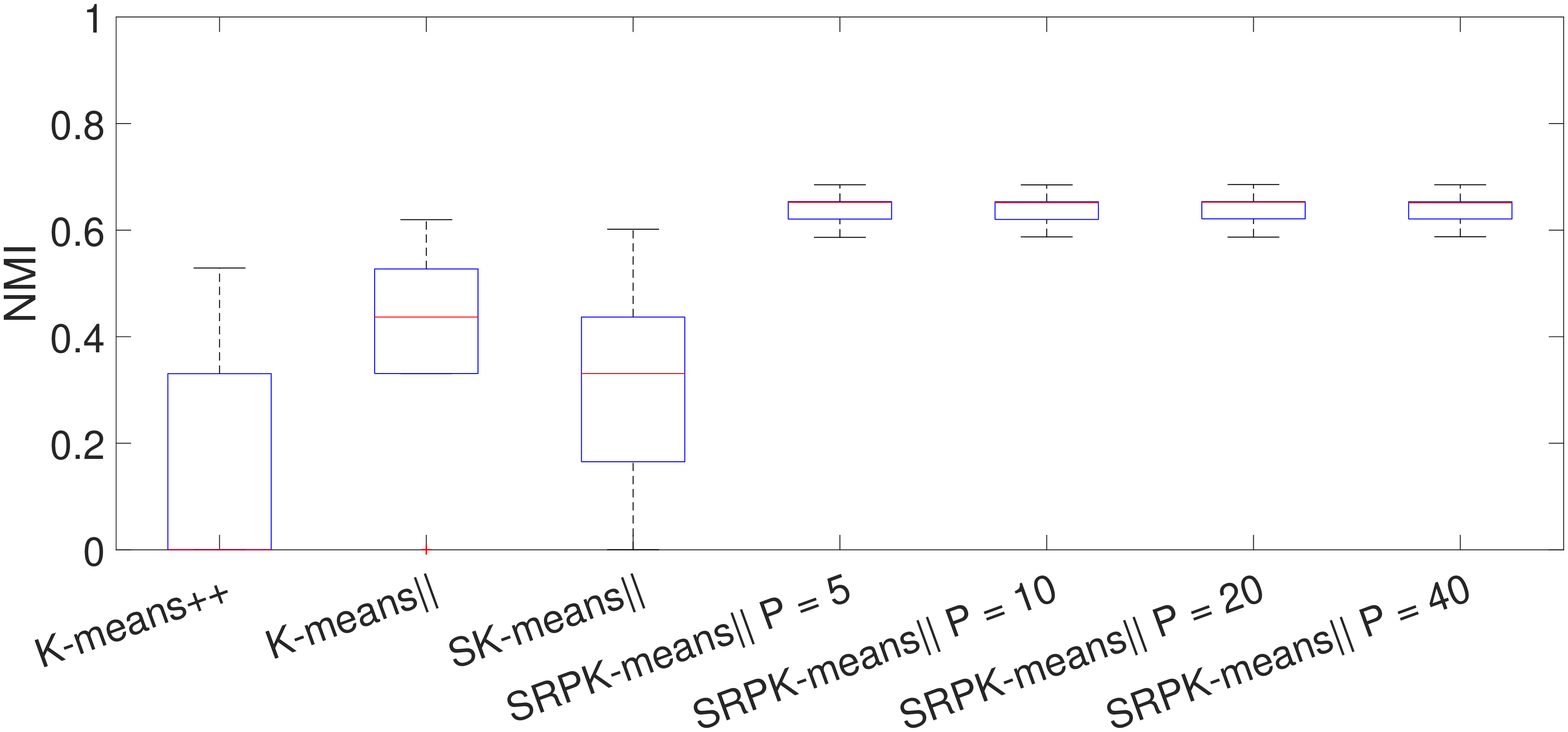}
    \caption{M-spheres-M1k-$d_c$0.05}
    \label{fig:kspheresM1kdc005}
  \end{subfigure} 
  ~
  \centering
  \begin{subfigure}{0.48\textwidth}
    \includegraphics[width=\textwidth]{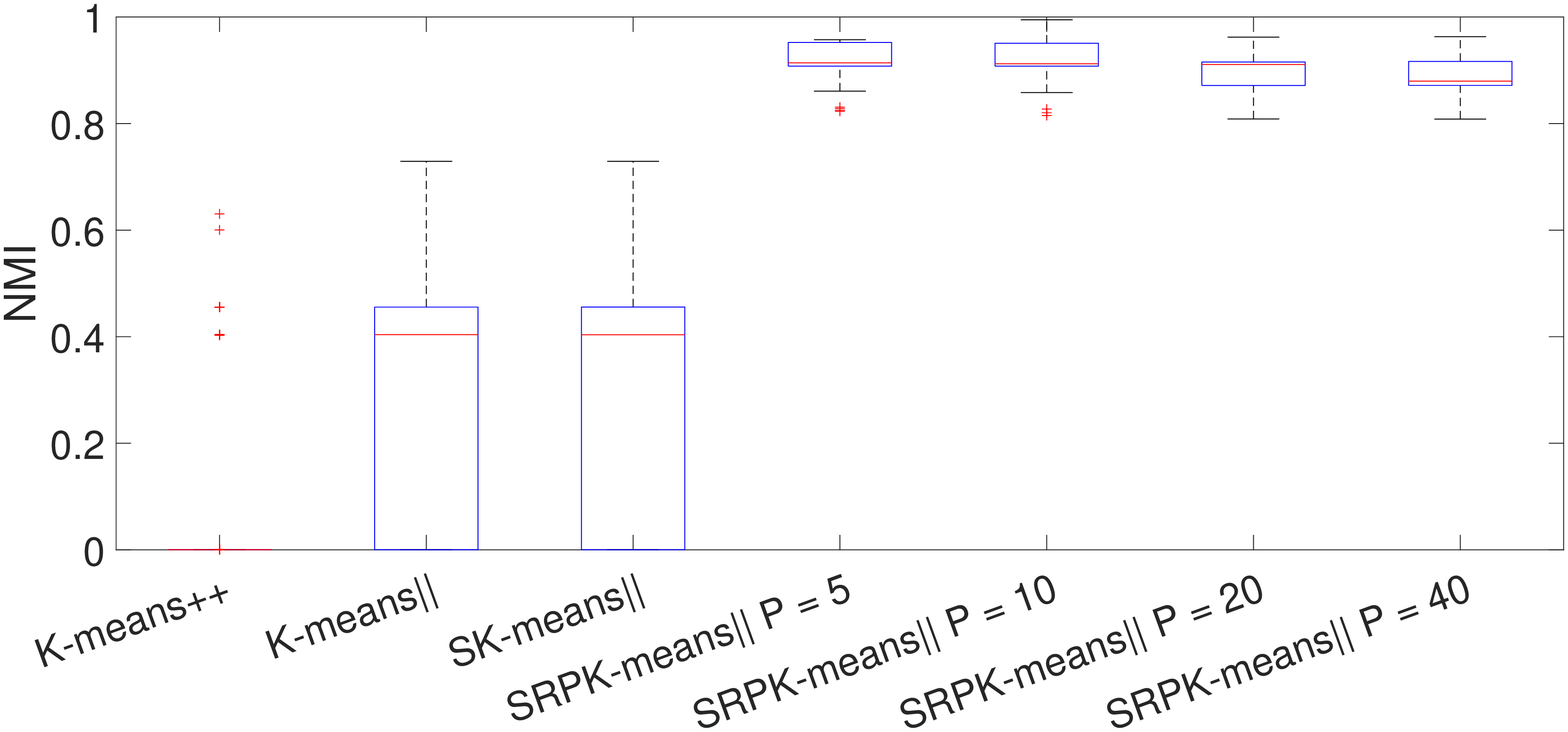}
    \caption{M-spheres-M10k-$d_c$0.05}
    \label{fig:kspheresM10kdc005}
  \end{subfigure} 
  \vspace{0.6cm}
  
  \centering
  \begin{subfigure}{0.48\textwidth}
    \includegraphics[width=\textwidth]{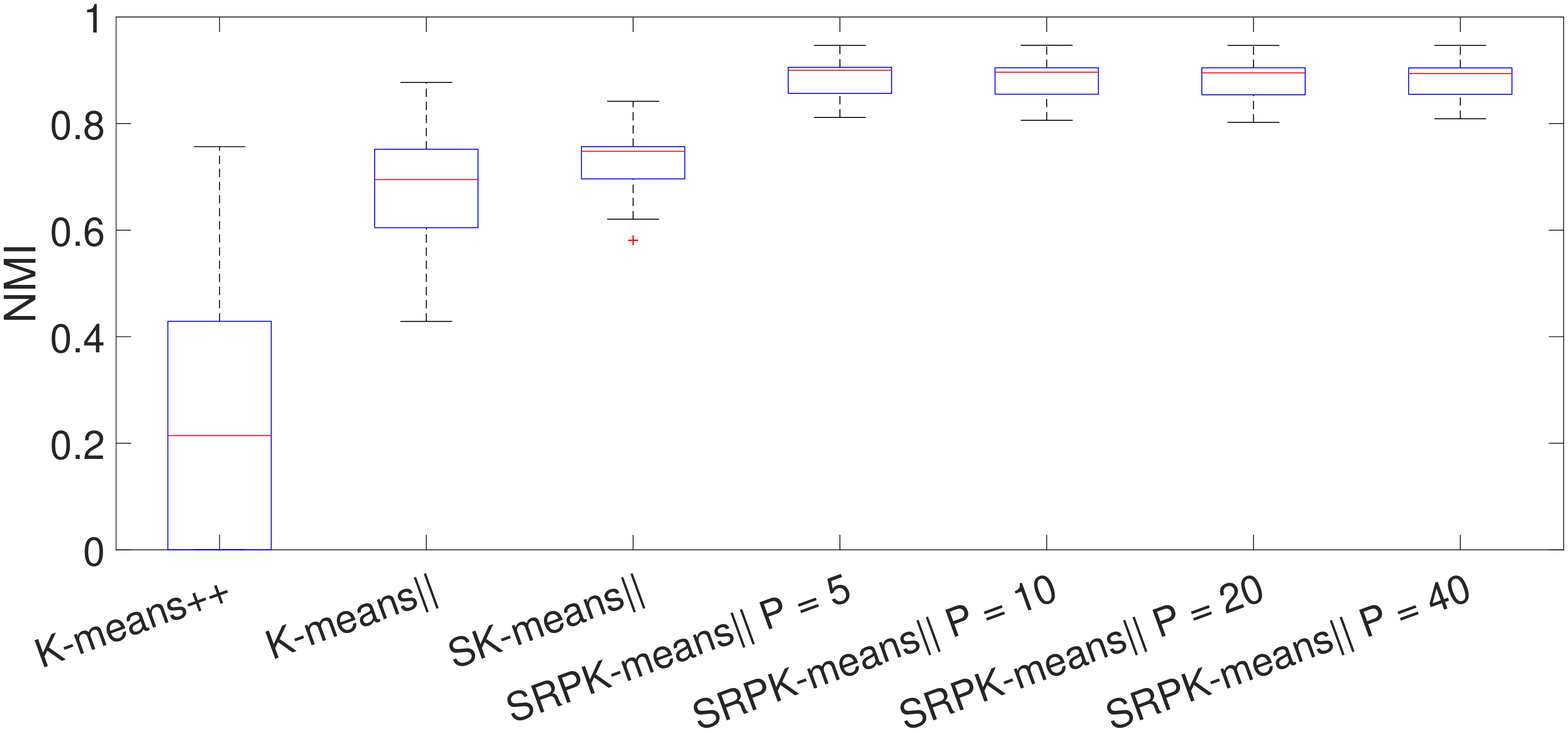}
    \caption{M-spheres-M1k-$d_c$0.1}
    \label{fig:kspheresM1kdc01}
  \end{subfigure} 
  ~
  \centering
  \begin{subfigure}{0.48\textwidth}
    \includegraphics[width=\textwidth]{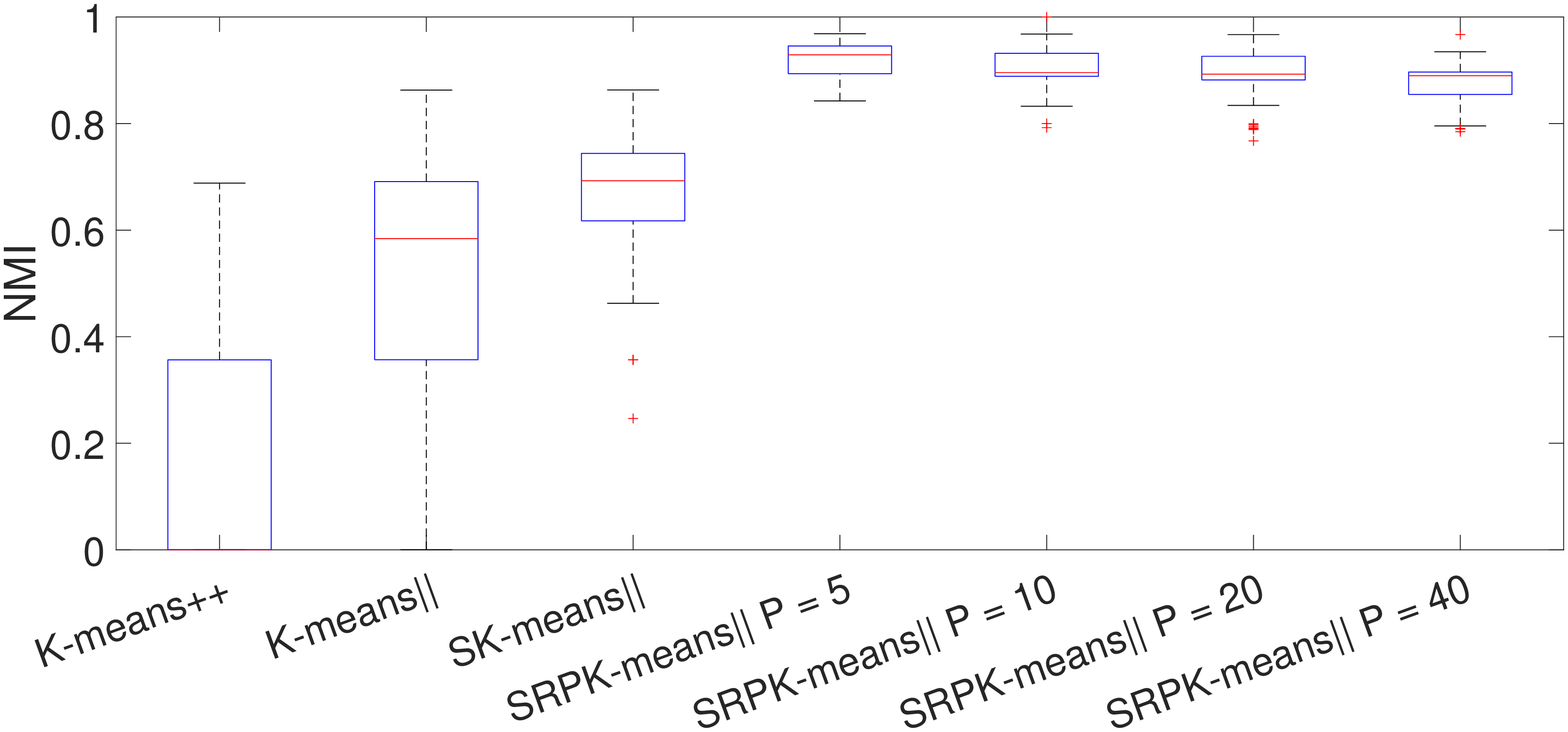}
    \caption{M-spheres-M10k-$d_c$0.1}
    \label{fig:kspheresM10kdc01}
  \end{subfigure} 
  \vspace{0.6cm} 
  
    \centering
  \begin{subfigure}{0.48\textwidth}
    \includegraphics[width=\textwidth]{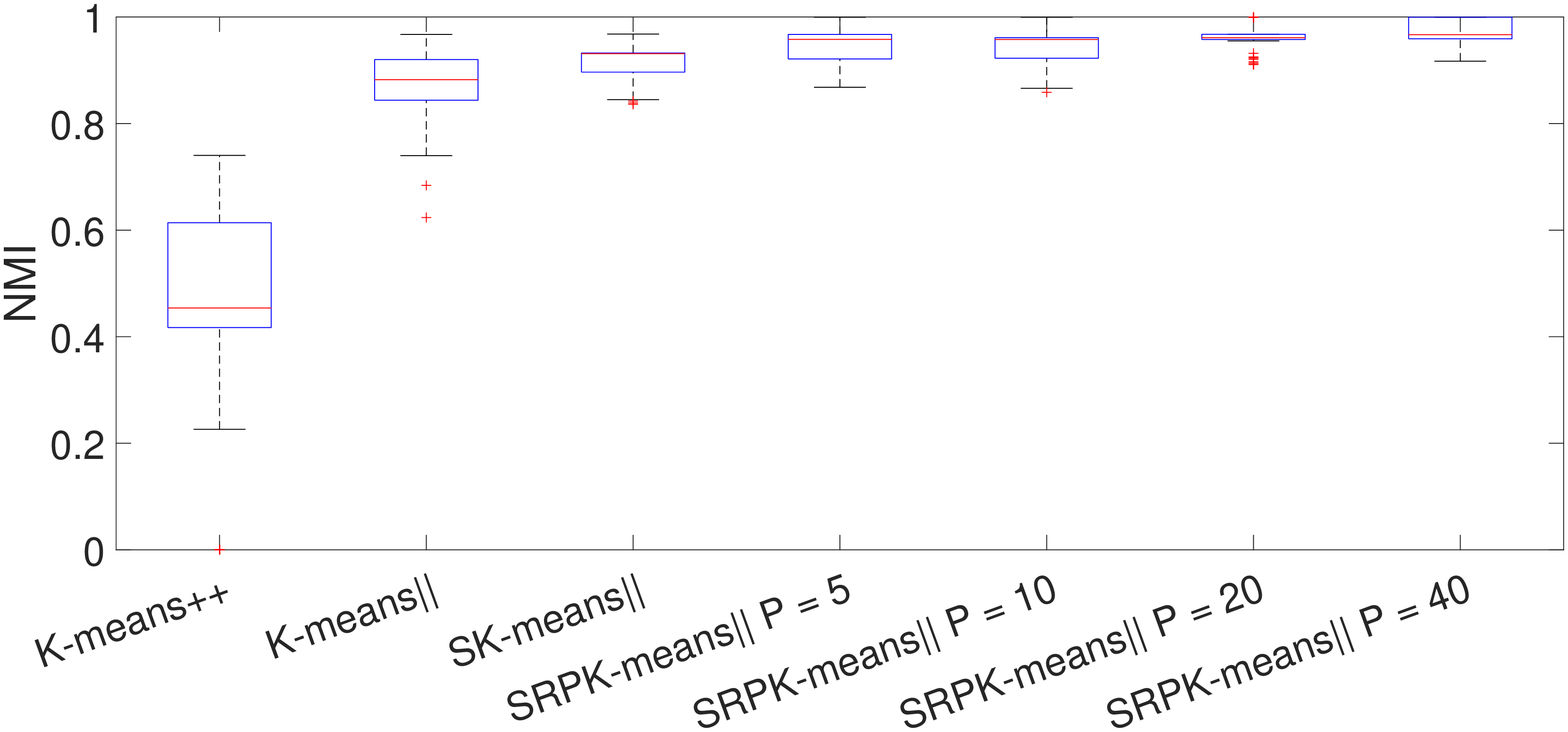}
    \caption{M-spheres-M1k-$d_c$0.2}
    \label{fig:kspheresM1kdc02}
  \end{subfigure} 
  ~
  \centering
  \begin{subfigure}{0.48\textwidth}
    \includegraphics[width=\textwidth]{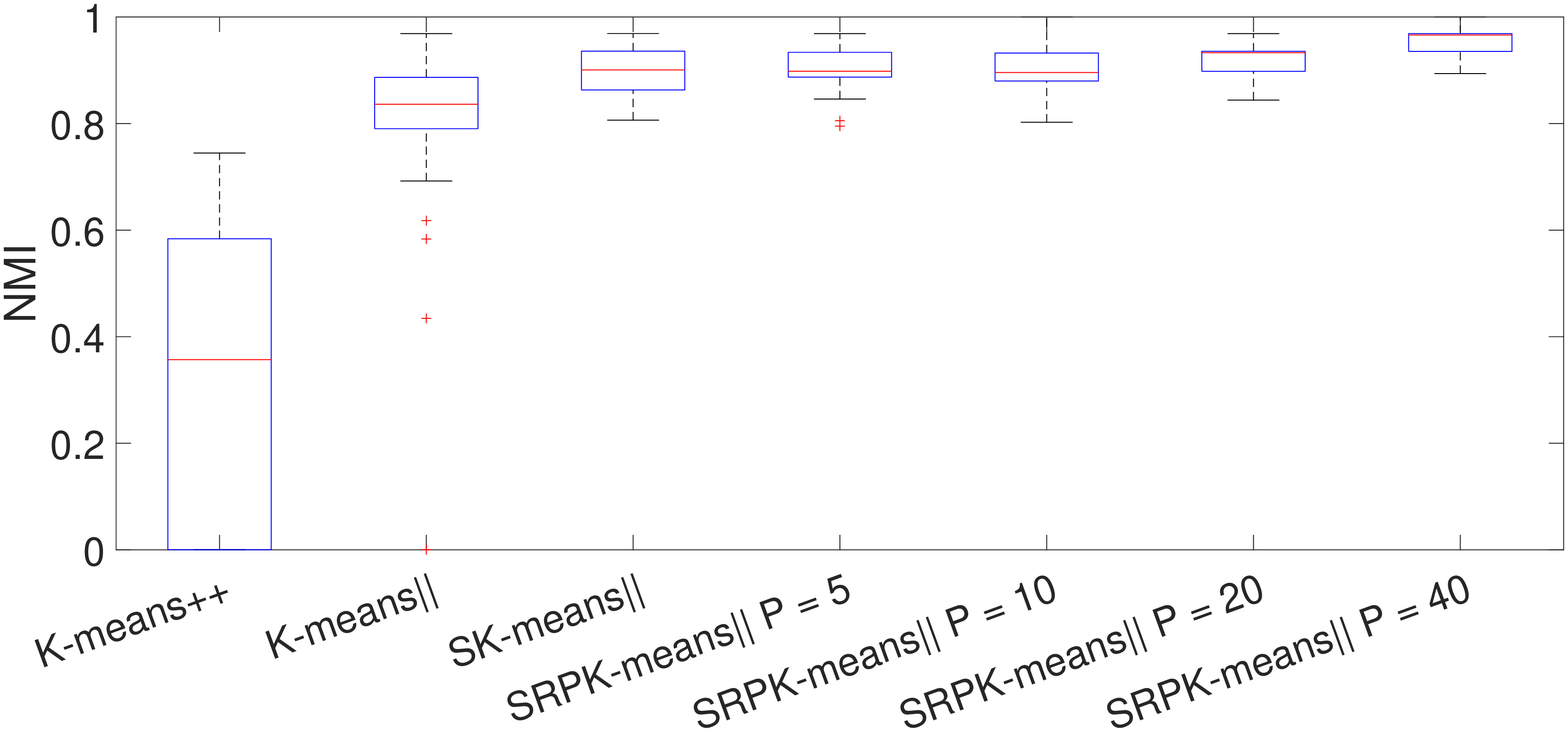}
    \caption{M-spheres-M10k-$d_c$0.2}
    \label{fig:kspheresM10kdc02}
  \end{subfigure} 
  \caption{NMI for the synthetic datasets.} 
  \label{fig:KSPHERES_NMI}
\end{figure*}

Currently, challenging simulated datasets for high-dimensional clustering problems are difficult to find. For instance, the experiments with DIM datasets of tens or hundreds of dimensions in \cite{hamalainen2017comparison} were inconclusive: all clustering results and cluster validation index comparisons behaved perfectly without any errors.

Therefore, we propose here a novel high-dimensional clustering dataset generation algorithm, where one can control the distances between the nearest cluster centers. The proposed algorithm generates $K$ nonuniform clusters where the points are within $M$-dimensional spheres in random locations. For simplicity and to standardize the comparisons, each cluster has an equal size and density radius





The proposed clustering dataset generator is given in Algorithm \ref{alg:gendata}. Based on the method proposed in \cite[p.~586]{muller1956some}, Algorithm \ref{alg:randsurfpoint} generates a random point that is on the $M$-dimensional sphere centered on $\c$ with radius of $d$. The core principle is to draw $M$ independent values from the standard normal distribution and transform these with corresponding $M$-direction cosines. The obtained $M$-dimensional vector is then scaled with the radius $d$ and relocated with the center $\c$. The generated points are uniformly distributed on the surface of the sphere, because of the known properties of the standard normal distribution (see \cite[p.~587]{muller1956some} and articles therein). The generator uses Algorithm \ref{alg:randsurfpoint} for generating $K$ cluster centers so that $\|\c_i-\c_j\| = d_c$, where $i \ne j$, $d_c$ is the given distance between the centers, and both $\c_i,\c_j$ then belong to the set of centers $\C$. Finally, $N_K$ data points for each cluster are generated by applying Algorithm \ref{alg:randsurfpoint} with a uniformly random radius from the interval $(0,d_r]$. This means, in particular, that points in a cluster are nonuniformly distributed and approximately $\frac{100a}{d_r} \%$ percentages of the points are within a $M$-dimensional sphere with radius $a$ for $0 \le a \le d_r$. In Algorithm \ref{alg:randsurfpoint}, $\mathcal{N}(0,1)$ denotes the standard normal distribution.


For simplicity, generation of the cluster centers starts from the origin. When the new centers are randomly located with the fixed distance and then expanded as clusterwise data in $\mathbb{R}^M$, the generator algorithm does not restrict the actual values of the generated data and centers. Hence, depending on the input, data range can be large. However, this is  alleviated with the min-max scaling as part of the clustering process. Details of the six generated datasets with Algorithm \ref{alg:gendata} are summarized in Table \ref{tab:synthdatasets}. The generated datasets are referred as M-spheres. For each dataset, we set $N_K = 10 000$, $K = 10$, and $d_r = 1$. To demonstrate interesting effects in the clustering initialization, we varied the nearest cluster center distance as $d_c = \{0.05,0.1,0.2\}$ and the data dimension as $M = \{1000,10000\}$. We set $d_c$ values to much smaller than $d_r$ in order to increase the difficulty of the clustering problems. In Figure \ref{fig:KSPHERES_PCA}, PCA projections on the three largest principal components show that the clusters are more separated for $M = 10 000$ than for $M = 1000$. For the M-spheres datasets, we used the serial implementations of the initialization methods with the same settings as before.

\begin{figure}[t]
    \centering
    \includegraphics[width=0.35\textwidth]{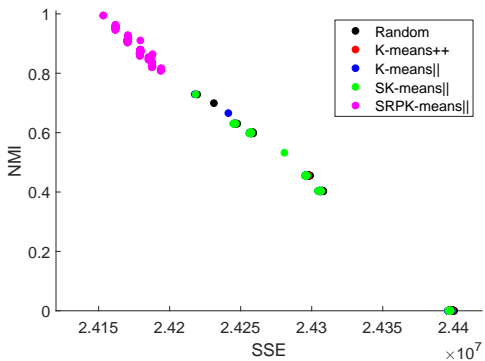}
  \caption{Scatter plot of SSE and NMI results for the M-spheres-M10k-$d_c$0.05 dataset.} 
  \label{fig:KSPHERES_Scatter}
\end{figure}

Results for the final clustering accuracy using NMI with $100$ repeats are shown in Figure \ref{fig:KSPHERES_NMI}. Clearly, SRPK-means$\parallel$ outperforms other methods in terms final clustering accuracy for all the synthetic datasets. Moreover, if we compare the results between the datasets with $M = 1000$ and $M = 10000$, we observe that the clustering accuracy for SRPK-means$\parallel$ is improved when the dimensionality increases. The most significant difference is obtained for the  M-spheres-M10k-$d_c$0.05 dataset, where K-means++ has a total breakdown of the accuracy while SRPK-means$\parallel$ is able find the near optimal clustering result out of $100$ repeats. Moreover, the accuracy of K-means++ is clearly worse compared to K-means$\parallel$ and SK-means$\parallel$ for very high dimensional datasets. We tested K-means with random initialization for this dataset and observed from the statistical testing that K-means++, K-means$\parallel$ and SK-means$\parallel$ are no better than the random initialization in terms of NMI. These results demonstrate that the use of distances in the K-means++ type of initialization strategies can become meaningless in very high-dimensional spaces.

In Figure \ref{fig:KSPHERES_Scatter}, a scatter plot of SSE and NMI values shows that the relative SSE difference between worst possible clustering result ($\text{NMI} = 0 $) and the optimal clustering result ($\text{NMI} = 1 $) can be surprisingly small for very high-dimensional data. Therefore, the improvements for the final clustering accuracy in Table \ref{tab:SSE} can be much more significant than the impression given by SSE in terms of how spherical clusters are found for high-dimensional datasets.

\section{Conclusion}

In this paper, we proposed two parallel initialization methods for large-scale K-means clustering and a new high-dimensional clustering data generation algorithm to support their empirical evaluation. The methods are based on divide-and-conquer type of K-means$\parallel$ approach and random projections. The proposed initialization methods are scalable and fairly easy to implement with different parallel programming models. 

The experimental results for 15 benchmark datasets showed that the proposed methods improve clustering accuracy and the speed of convergence compared to state-of-the-art approaches. Experiments with SRPK-means$\parallel$ method demonstrate that utilization of RP and K-means$\parallel$ is beneficial for clustering large-scale high-dimensional datasets. In particular, SRPK-means$\parallel$ is an appealing approach as a standalone algorithm for clustering very high-dimensional large-scale datasets. The confirmed finding (e.g., \cite{aggarwal2001surprising}) that the difference between the errors (SSE) of good and bad clustering results in high-dimensional spaces can be surprisingly small also challenge cluster validation and cluster validation indices (see \cite{hamalainen2017comparison} and references therein) in such cases. 

Figures \ref{fig:KSPHERES_NMI} and \ref{fig:KSPHERES_Scatter} illustrate the deteriorating behavior of the currently most popular K-means++ initialization method in high dimensions. We especially observe that the K-means++ initialization behaves like (i.e., is not better than) the random one in the very high-dimensional cases. Such finding also suggests further experiments, where as a function of the data dimension, emergence of such a behavior is being studied to identify most appropriate random project dimensions to restore the quality of initialization and the whole clustering algorithm.



%



\ifCLASSOPTIONcompsoc
  \section*{Acknowledgments}
\else
  \section*{Acknowledgment}
\fi

The work has been supported by the Academy of Finland from the projects 311877 (Demo) and 315550 (HNP-AI).

\ifCLASSOPTIONcaptionsoff
  \newpage
\fi

\bibliographystyle{IEEEtran}
\bibliography{references}

\end{document}